\documentclass{article}

\PassOptionsToPackage{numbers}{natbib}

     \usepackage[final]{neurips_2021}

\usepackage[utf8]{inputenc} 
\usepackage[T1]{fontenc}    
\usepackage{hyperref}       
\usepackage{url}            
\usepackage{booktabs}       
\usepackage{amsfonts}       
\usepackage{nicefrac}       
\usepackage{microtype}      
\usepackage{xcolor}         
\usepackage{multirow}

\usepackage{amsmath,amsfonts,bm}

{}
{}
{}
{}

\def\eqref#1{equation~\ref{#1}}

\def\1{\bm{1}}

\def\vtheta{{\bm{\theta}}}

\def\vc{{\bm{c}}}

\def\vf{{\bm{f}}}

\def\vh{{\bm{h}}}

\def\vk{{\bm{k}}}

\def\vo{{\bm{o}}}

\def\vq{{\bm{q}}}

\def\vu{{\bm{u}}}
\def\vv{{\bm{v}}}

\def\vx{{\bm{x}}}
\def\vy{{\bm{y}}}
\def\vz{{\bm{z}}}

\def\mR{{\bm{R}}}

\def\mW{{\bm{W}}}

\DeclareMathAlphabet{\mathsfit}{\encodingdefault}{\sfdefault}{m}{sl}
\SetMathAlphabet{\mathsfit}{bold}{\encodingdefault}{\sfdefault}{bx}{n}

\newcommand{\softmax}{\mathrm{softmax}}

\newcommand{\SlowNet}{\mathtt{SlowNet}}
\newcommand{\FastNet}{\mathtt{FastNet}}

\newcommand{\UpdateRule}{\mathtt{UpdateRule}}
\newcommand{\SlowSubnet}{\mathtt{SlowSubnet}}
\newcommand{\sigmoid}{\sigma}

\usepackage{graphicx}
\newcommand{\mpm}[2]{#1 $\pm$ {\small #2}}  

\title{Going Beyond Linear Transformers \\with Recurrent Fast Weight Programmers}
\author{%
  Kazuki Irie$^{1}$\footnotemark[1] , Imanol Schlag$^{1}$\thanks{Equal contribution.} , R\'obert Csord\'as$^{1}$, J\"urgen Schmidhuber$^{1,2}$\\
  $^1$The Swiss AI Lab, IDSIA, University of Lugano (USI) \& SUPSI, Lugano, Switzerland \\
  $^2$King Abdullah University of Science and Technology (KAUST), Thuwal, Saudi Arabia \\
  \texttt{\{kazuki, imanol, robert, juergen\}@idsia.ch} \\
}

\begin{document}

\maketitle

\begin{abstract}
Transformers with linearised attention (``linear Transformers'') have demonstrated the practical scalability and effectiveness of outer product-based Fast Weight Programmers (FWPs) from the~'90s.
However, the original FWP formulation is more general than the one of linear Transformers: a \textit{slow} neural network (NN) continually reprograms the weights of a \textit{fast} NN with \textit{arbitrary}  architecture.
In existing linear Transformers, both NNs are feedforward and consist of a single layer.
Here we explore new variations by adding recurrence to the slow and fast nets.
We evaluate our novel recurrent FWPs (RFWPs) on two synthetic algorithmic tasks (code execution and sequential ListOps), Wikitext-103 language models, and on the Atari 2600 2D game environment.
Our models exhibit properties of Transformers and RNNs. 
In the reinforcement learning setting, we report large improvements over LSTM in several Atari games.
Our code is public.\footnote[1]{\url{https://github.com/IDSIA/recurrent-fwp}}
\end{abstract}

\section{Introduction}
\label{sec:intro}
The Transformer \citep{trafo} has become one of the most popular neural networks (NNs) for processing sequential data.
Its success on neural machine translation quickly transferred to other problems in natural language processing, such as language modelling \citep{al2018character, gpt3} or question answering \citep{devlin2019bert}.
Recently, it has also been applied in other domains, including image processing \citep{dosovitskiy2021an, zhu2021deformable} or mathematical problem solving \citep{saxton2018analysing, schlag2019enhancing, charton2021learning}.

Conceptually, the Transformer is a deep feedforward NN that processes all elements of a  sequence in parallel:
unlike in recurrent NNs (RNNs), the computations of a layer for the entire sequence can be packed into one big matrix multiplication.
This scales well with the number of parallel processors.

Despite the benefits of parallelisation, a major drawback of Transformers
is that their computational complexity in time and space is quadratic in sequence length.
Furthermore, in the auto-regressive version \citep{trafo, al2018character} --- the focus of our work ---
the state size increases linearly with sequence length.
This makes Transformers infeasible for auto-regressive settings dealing with very long or potentially infinite sequences, forcing practitioners to truncate temporal contexts and ignore long-term  dependencies beyond fixed-size time windows.
Although recent work tries to address this issue \citep{dai2019transformerxlacl, rae2020compressive},
this limitation makes some applications of Transformers challenging, e.g.,
reinforcement learning (RL) in partially observable environments \citep{parisotto2020stabilizing, parisotto2021efficient}, which is still dominated by RNNs such as the Long Short-Term Memory (LSTM; \cite{hochreiter1997long}) trained by policy gradients \cite{wierstra2010, wierstra2007, openai2019dota, vinyals2019grandmaster}. 

To scale Transformers to longer sequences, recent works have proposed to linearise the softmax in the self-attention computation
and reorganise the latter in a sequential way \citep{katharopoulos2020transformers}.
Such models include \citeauthor{katharopoulos2020transformers}'s \textit{Linear Transformer} (LT) \citep{katharopoulos2020transformers}, \citeauthor{choromanski2020rethinking}'s \textit{Performer} \citep{choromanski2020rethinking}
and \citet{peng2021random}'s variant.
They enjoy time and space complexities linear in sequence length with states of constant size.
While their performance on some tasks does not fully match the one of regular Transformers \citep{tay2020long}, several improvements have already been proposed \citep{peng2021random, schlag2021linear} (see our review in Sec.~\ref{sec:examples}) which makes this Transformer family a promising alternative.

Here we go one step further in advancing linear Transformer variants as powerful auto-regressive sequence processing models,
adopting the perspective of ``Fast Weight Programmers'' (FWPs) \citep{Schmidhuber:91fastweights, schmidhuber1992learning, schmidhuber1993reducing}.
Recent work emphasised that linearised Transformers are essentially equivalent to outer product-based FWPs from the~'90s (\citep{schlag2021linear}; reviewed in Sec.~\ref{sec:background}).
Here we explore this connection further and describe more powerful FWPs. 

The original FWP \citep{Schmidhuber:91fastweights} is a two-NN system: a slow and a fast net, each with arbitrary architectures. The slow net learns to generate rapid context-dependent weight modifications for the fast net. 
In the case of existing linear Transformer variants, the slow and fast nets are simple one layer feedforward NNs. 
Here we augment them with recurrent connections to obtain recurrent FWPs (RFWPs).
Recurrence enhances the model's theoretical power \citep{hahn2020theoretical} and
can help to solve tasks that naturally require
recurrence as a part of the solution.

Our experiments on the language modelling dataset Wikitext-103 \citep{merity2016pointer} show that our RFWPs are competitive compared to regular Transformers.
We then study various properties of the proposed models on two synthetic algorithmic tasks: code execution \cite{zaremba2014learning} and sequential ListOps \cite{nangia-bowman-2018-listops}.
Finally, it is straightforward to apply our models to RL problems as a drop-in replacement for LSTMs.
Here our RFWPs obtain large improvements over LSTM baselines across many Atari 2600 2D game environments  \citep{BellemareOGTM16}.
Although LSTM still works better in a few environments, we show  that our RFWPs generally improve by scaling them up.

The main contribution of this work is twofold:
(1) from the perspective of FWPs,
we study novel powerful
FWPs for sequence processing,
demonstrating that NNs can easily learn to control
NNs that are more complex than a single feedforward layer,
and (2) from the perspective of Transformer models,
our RFWPs augment linear Transformers
with recurrence, addressing general limitations of existing
auto-regressive Transformer models.

\section{Background on Fast Weight Programmers (FWPs)}
\label{sec:background}
Here we review the general concept of  FWPs,
as well as two specific instances thereof:
the linear Transformer \citep{katharopoulos2020transformers, choromanski2020rethinking} and the 
Delta Net \citep{schlag2021linear}.

\subsection{General Formulation}
\label{sec:general}
We refresh the concept of fast weight controllers or FWPs \citep{Schmidhuber:91fastweights, schmidhuber1992learning} using modern notation in
a sequence processing scenario.
An FWP with trainable parameters $\vtheta_{\text{slow}}$ sequentially transforms an input sequence $\{ \vx^{(t)} \}_{t=1}^T$ with $\vx^{(t)} \in \mathbb{R}^{d_\text{in}}$ to an output sequence
$\{ \vy^{(t)} \}_{t=1}^T$ with $\vy^{(t)} \in \mathbb{R}^{d_\text{out}}$ of length $T$ as
\begin{align}
\vtheta_{\text{fast}}^{(t)}, \vq^{(t)} &= \SlowNet\big(\{\vx^{(j)}\}_{j=1}^t, \{\vy^{(j)}\}_{j=0}^{t-1}, \{\vtheta^{(j)}_{\text{fast}}\}_{j=0}^{t-1},
\{\vq^{(j)}\}_{j=0}^{t-1}; \vtheta_{\text{slow}}\big) \label{eq:slow_net} \\
\vy^{(t)} &= \FastNet(\{\vq^{(j)}\}_{j=1}^t, \{\vy^{(j)}\}_{j=0}^{t-1} ; \vtheta^{(t)}_{\text{fast}})
\label{eq:fast_net}
\end{align}
where $\vy^{(0)}$, $\vtheta^{(0)}_{\text{fast}}$, and $\vq^{(0)}$ are initial variables.
This is a system with two NNs called $\FastNet$ and $\SlowNet$
in which
the parameters $\vtheta_{\text{fast}}^{(t)}$ of $\FastNet$ are generated by  $\SlowNet$ at each time step $t$.
The weights of the fast net are \textit{fast} in the sense that they may rapidly change at
every step of the sequence while the weights of the slow net $\vtheta_{\text{slow}}$ are
\textit{slow} because they can only change through gradient descent during training, remaining fixed afterwards\footnote{The fast net could also contain some additional slow weights; we omit this possibility here.}.
Eq.~\ref{eq:slow_net} expresses a slow NN in its general form. 
The slow net can generate fast weights
conditioned on various variables, depending on architectural choices for the slow and fast NNs.
In addition to the fast weights $\vtheta_{\text{fast}}^{(t)}$, the slow net also generates or \textit{invents} an input $\vq^{(t)}$ to be fed to the fast net (alternatively $\vq^{(t)}$ can simply be $\vx^{(t)}$). 
While the architectures of slow and fast nets are arbitrary, they are
typically chosen to be differentiable such that the entire FWP can be trained in an end-to-end manner using gradient descent. 
By interpreting the weights of an NN as a program \citep{Schmidhuber:90diff}, the slow net effectively learns to control, or \textit{program}, the fast NN. 
Thus, the slow net is a neural programmer of fast weights,
and its parameter set $\vtheta_{\text{slow}}$ embodies compressed 
information used to produce potentially infinite variations of
context-dependent fast weights.

In many settings, it makes sense to generate the fast weights $\vtheta_{\text{fast}}^{(t)}$ incrementally in an iterative fashion, where  the $\SlowNet$ is further decomposed into two sub-parts:
\begin{align}
\label{eq:subslow}
\vz^{(t)}, \vq^{(t)} &= \SlowSubnet(
\{\vx^{(j)}\}_{j=1}^t, \{\vy^{(j)}\}_{j=0}^{t-1}, \{\vtheta^{(j)}_{\text{fast}}\}_{j=0}^{t-1},
\{\vq^{(j)}\}_{j=0}^{t-1},
\{\vz^{(j)}\}_{j=0}^{t-1}; \vtheta_{\text{slow}}) \\
\vtheta^{(t)}_{\text{fast}} &= \UpdateRule(\vtheta^{(t-1)}_{\text{fast}}, \vz^{(t)})
\end{align}
where $\UpdateRule$ takes the fast weights $\vtheta^{(t-1)}_{\text{fast}}$ from the previous iteration to produce the new fast weights $\vtheta^{(t)}_{\text{fast}}$ conditioned on $\vz^{(t)}$. 
The update rule is essentially the differentiable \textit{elementary programming instruction} used by the FWP.
In the next section we review concrete examples of recent FWPs.

\subsection{Linear Transformers as Fast Weight Programmers}
\label{sec:examples}
In general, the dimension of the fast weights $\vtheta^{(t)}_{\text{fast}}$ is too large to be conveniently parameterised by an NN. 
Instead, it was proposed in 1991 \citep{Schmidhuber:91fastweights} 
to perform a rank-one update via the outer product of two vectors generated by the slow net.
Two recent models directly correspond to such outer product-based FWPs: linear Transformers \citep{katharopoulos2020transformers} and the Delta Net \citep{schlag2021linear}.

\paragraph{Linear Transformer.}
The ``linear Transformer'' \citep{katharopoulos2020transformers} is a class of Transformers where the softmax in the attention is linearised.
This is achieved by replacing the softmax with a kernel function $\phi$---then the self-attention can be rewritten as a basic outer product-based FWP \citep{Schmidhuber:91fastweights, schlag2021linear}. 
Previous works focused on different $\phi$ maps with properties such as increased capacity \citep{schlag2021linear} or guaranteed approximation of the softmax in the limit \citep{choromanski2020rethinking, peng2021random}.
For our purposes, the particular choice of $\phi$ is irrelevant and we simply assume $\phi: \mathbb{R}^{d_\text{key}} \rightarrow \mathbb{R}^{d_\text{key}}$, simplifying our equations below by writing $\vk, \vq$ instead of $\phi(\vk), \phi(\vq)$.
Using otherwise the same notation as above, for each new input $\vx^{(t)}$, the output $\vy^{(t)}$
is obtained by:
\begin{eqnarray}
\label{eq:proj}
\vk^{(t)}, \vv^{(t)}, \vq^{(t)} &=& \mW_k \vx^{(t)}, \mW_v \vx^{(t)}, \mW_q \vx^{(t)} \\
\mW^{(t)}  &=& \mW^{(t-1)} + \vv^{(t)}  \otimes \vk^{(t)} \label{eq:fw_add} \\
\vy^{(t)}  &=& \mW^{(t)} \vq^{(t)} \label{eq:fw_get}
\end{eqnarray}
where the slow weight matrices $\mW_k \in \mathbb{R}^{d_\text{key} \times d_\text{in}}$ and $\mW_v \in  \mathbb{R}^{d_\text{out} \times d_\text{in}}$ are used to obtain the \textit{key} $\vk^{(t)} \in \mathbb{R}^{d_\text{key}}$ and the \textit{value} $\vv^{(t)} \in \mathbb{R}^{d_\text{out}}$.
The key and value vectors are used to generate new weights via the outer product $\vv^{(t)} \otimes \vk^{(t)} \in \mathbb{R}^{d_\text{out} \times d_\text{key}}$.
A further simplification in the equations above is the omission of attention normalisation which has been experimentally shown to be unnecessary if the $\phi$ function produces normalised key and query vectors \citep{schlag2021linear}.

In Eq.~\ref{eq:fw_add}, the previous fast weight matrix $\mW^{(t-1)} \in \mathbb{R}^{d_\text{out} \times d_\text{key}}$ is updated to yield $\mW^{(t)}$ by adding the update term $\vv^{(t)}  \otimes \vk^{(t)}$.
This corresponds to the \textit{sum update rule}
or \textit{purely additive programming instruction}.
Here the fast NN is a simple linear transformation as in Eq.~\ref{eq:fw_get} which takes as input the query vector $\vq^{(t)} \in \mathbb{R}^{d_\text{key}}$ generated by the slow weights $\mW_q \in \mathbb{R}^{d_\text{key} \times d_\text{in}}$.
Hence, in linear Transformers, the previous Eq.~\ref{eq:subslow} simplifies to: $\vz^{(t)}, \vq^{(t)} = \SlowSubnet(\vx^{(t)}; \vtheta_{\text{slow}})$ with $\vz^{(t)}= (\vk^{(t)}, \vv^{(t)})$.

\paragraph{Delta Net.}
The Delta Net \citep{schlag2021linear} is obtained by replacing the purely additive programming instruction (Eq.~\ref{eq:fw_add}) in the linear Transformer with the one akin to the \textit{delta rule} \citep{widrow1960adaptive}:
\begin{align}
\label{eq:delta}
\mW^{(t)} &= \mW^{(t-1)} + \beta^{(t)}(\vv^{(t)} - \bar{\vv}^{(t)}) \otimes \vk^{(t)} 
\end{align}
where $\beta^{(t)} \in \mathbb{R}$ is a fast parameter (learning rate) of the update rule generated by the slow net with weights $\mW_\beta \in \mathbb{R}^{1 \times d_\text{in}}$ and the sigmoid function $\sigmoid$:
\begin{align}
\label{eq:beta}
\beta^{(t)} &= \sigmoid(\mW_\beta \vx^{(t)})
\end{align}
and $\bar{\vv}^{(t)}  \in \mathbb{R}^{d_\text{out}}$ is generated as a function of the previous fast weights $\mW^{(t-1)}$ and the key $ \vk^{(t)}$
\begin{align}
\bar{\vv}^{(t)} &= \mW^{(t-1)} \vk^{(t)}.
\end{align}
This update rule was introduced to address a memory capacity problem affecting linear Transformers with the purely additive update rule \citep{schlag2021linear}.
The corresponding Eq.~\ref{eq:subslow} is: $\vz^{(t)}, \vq^{(t)} = \SlowSubnet(\vx^{(t)}, \mW^{(t-1)}; \vtheta_{\text{slow}})$ with $\vz^{(t)}= (\vk^{(t)}, \vv^{(t)}, \beta^{(t)}, \bar{\vv}^{(t)})$.
Thus, unlike linear Transformers, the $\SlowNet$ in the Delta Net
takes the previous fast weights $\mW^{(t-1)}$ into account to generate the new fast weight updates.

We typically use the multi-head version \citep{trafo} of the computations above.
After the projection (Eq.~\ref{eq:proj}), the vectors $\vk^{(t)}$, $\vv^{(t)}$, $\vq^{(t)}$ are split into equally sized $H$ sub-vectors, and the rest of the operations
are conducted by $H$ computational heads independently.
The resulting output vectors from each head are concatenated
to form the final output.

\paragraph{Other approaches.}
While our focus here is on outer product-based weight generation, which is an efficient method to handle high dimensional NN weights, there are also other approaches.
For example, instead of generating a new weight matrix, Hypernetworks \citep{ha2017hypernetworks} scale the rows of a slow weight matrix with a generated vector of appropriate size.
Weight compression to control fast weights in a low dimensional compressed space has been also studied \citep{irie2021training}.
In the broad sense of context-dependent weights \citep{von1981correlation, feldman1982dynamic, mcclelland1985putting}, 
many concepts relate to FWPs:  e.g.~dynamic convolution
\citep{KleinWA15, noh2016image, NIPS2016_8bf1211f},
LambdaNetworks \citep{bello2021lambdanetworks},
or dynamic plasticity \citep{pmlr-v80-miconi18a, miconi2018backpropamine}.

\section{Fast Weight Programmers With Slow or Fast RNNs}
\label{sec:model}

The original formulation of FWPs reviewed in Sec.~\ref{sec:general} is more general
than existing models presented in Sec.~\ref{sec:examples}.
In particular, both fast and slow networks in existing linear Transformers consist of a single feedforward layer (Eqs.~\ref{eq:proj} and \ref{eq:fw_get}).
Here we present FWPs with recurrent fast nets in Sec.~\ref{sec:fast-net} and FWPs with recurrent slow nets in Sec.~\ref{sec:slow-net}.

\subsection{Fast Network Extensions}
\label{sec:fast-net}

In principle, any NN architecture can be made \textit{fast}.
Its fast weight version is obtained by replacing the networks' weights with fast weights parameterised by an additional slow network.
For example, consider a regular RNN layer with two weight matrices $\mW$ and $\mR$: 
\begin{eqnarray}
\label{eq:basic_rnn}
\vh^{(t)}  &=& \sigma(\mW \vx^{(t)} + \mR \vh^{(t-1)})
\end{eqnarray}
A fast weight version can be obtained by replacing $\mW$ and $\mR$ with $\mW^{(t)}$ and $\mR^{(t)}$
which are controlled as in Eq.~\ref{eq:delta} with all necessary variables
generated by a separate slow net at each time step $t$.

While this view illustrates the generality of FWPs,
the angle under which we approach these models is slightly different:
we introduce recurrence as a way of augmenting existing linear Transformers.

\paragraph{Delta RNN.}
We obtain a fast weight RNN called \textbf{Delta RNN}  by adding an additional recurrent term to the feedforward fast net of the linear Transformer (Eq.~\ref{eq:fw_get}):
\begin{eqnarray}
\label{eq:fastrnn}
\vy^{(t)}  &=& \mW^{(t)} \vq^{(t)} + \mR^{(t)} f(\vy^{(t-1)})
\end{eqnarray}
where $\mR^{(t)} \in \mathbb{R}^{d_\text{out} \times d_\text{out}}$ is an additional fast weight matrix which introduces recurrent connections.
It is also generated by the slow net using the delta update rule, similar to $\mW^{(t)}$ in Eq.~\ref{eq:delta} but with additional slow weights.
We apply an element-wise activation function $f$ to  the previous output of the fast network $\vy^{(t-1)}$ to obtain the recurrent query.
The choice of activation function is crucial here because, to achieve stable model behaviour, the elements in key and query vectors should be positive and sum up to one when the delta update rule is used \citep{schlag2021linear}.
We use the softmax function ($f = \softmax$ in Eq.~\ref{eq:fastrnn}) to satisfy these conditions.
An ablation study on the choice of using Eq.~\ref{eq:fastrnn} instead of the one similar to Eq.~\ref{eq:basic_rnn} can be found in Appendix \ref{app:lm_models}.

Analogous to the Delta RNN, we also construct a \textbf{Delta LSTM} with six fast weight matrices.
The exact equations can be found in Appendix \ref{app:lm_models}. 

\paragraph{Alternative Feedforward Fast Nets.}
While the focus of this work is on RNNs,
there are also interesting fast feedforward models to be used in Eq.~\ref{eq:fw_get}
which might result in stronger feedforward baselines.
For example, we can replace the single layer fast net of Eq.~\ref{eq:fw_get}
by a $K$-layer deep network:
\begin{eqnarray}
\vh_k^{(t)}  &=& \mW_k^{(t)} f(\vh_{k-1}^{(t)}) \quad  \text{for} \enspace k \in [1..K] 
 \enspace \text{with}  \enspace \vh_{0}^{(t)} = \vq^{(t)} \\
\vy^{(t)} &=& \vh_K^{(t)}  
\end{eqnarray}
where the slow network produces all $K$ fast weights $\{\mW_k^{(t)}\}_{k=1}^{K}$ and query $\vq^{(t)}$ from a single input $\vx^{(t)}$.
In light of the capacity limitation in linear Transformers \citep{schlag2021linear}, this might introduce additional capacity without the need of larger representations, analogous to the trade-off in a multilayer perceptron (MLP) between narrow \& deep versus shallow \& wide.
We refer to this class of models as \textbf{Delta MLPs}.
Again, for stable model behaviour with the delta rule, we apply the softmax activation $f$ to the vectors to be used as a query. 

Another interesting approach is to use a Delta Net itself as a fast net,
i.e., make the slow weights in the Delta Net fast (thus obtaining a \textbf{Delta Delta Net}).
Such a model could in principle learn to adapt the way of generating
fast weights depending on the context.
While we plan to investigate the potential of such hierarchical FWPs in future work, we also include preliminary results of such a model in our language modelling experiments (Sec.~\ref{sec:lm}).
A discussion on the dimensionality of such a
model can also be found in Appendix \ref{app:deltadelta}.

We experimentally demonstrate that (slow) NNs can learn to control the weights of these rather complex fast networks (Sec.~\ref{sec:exp}).

\subsection{Slow Network Extensions}
\label{sec:slow-net}
In linear Transformers, the slow network is purely feedforward (Eq.~\ref{eq:proj}).
It can be made recurrent at two different levels:
within the slow network (i.e.~the slow network computes 
weight updates based on its own previous outputs e.g., key, value, query vectors) or via the fast network by taking the fast net's previous output as an input.
In our preliminary experiments, we found the former to be sub-optimal (at least in language modelling experiments).
So we focus on the latter approach: we make the slow net in the Delta Net dependent on the previous output of the fast network.
We refer to this model as the \textbf{Recurrent Delta Net} (RDN).

\paragraph{Recurrent Delta Net.}
We obtain the RDN by modifying the generation of key, value, and query vectors (Eq.~\ref{eq:proj}) as well as the learning rate (Eq.~\ref{eq:beta}) in the Delta Net.
We add additional slow weights ($\mR_k, \mR_q \in \mathbb{R}^{d_\text{key} \times d_\text{out}}$, $\mR_v \in \mathbb{R}^{d_\text{out} \times d_\text{out}}$, and $\mR_\beta \in \mathbb{R}^{1 \times d_\text{out}}$) for recurrent connections
which connect the previous output of the fast net $\vy^{(t-1)}$ (Eq.~\ref{eq:fw_get}) to
the new $\vk^{(t)}$, $\vv^{(t)}$, $\vq^{(t)}$, and $\beta^{(t)}$ as follows:
\begin{eqnarray}
\vk^{(t)} &=& \mW_k \vx^{(t)} +  \mR_k \tanh(\vy^{(t-1)}) \\
\vv^{(t)} &=& \mW_v \vx^{(t)} +  \mR_v \tanh(\vy^{(t-1)}) \\
\vq^{(t)} &=& \mW_q \vx^{(t)} +  \mR_q \tanh(\vy^{(t-1)}) \\
\beta^{(t)} &=& \sigmoid(\mW_\beta \vx^{(t)} + \mR_\beta \tanh(\vy^{(t-1)}))
\end{eqnarray}
While the rest of the model remains as in the Delta Net,
with these simple extra recurrent connections the model becomes a proper RNN.
The corresponding dependencies in Eq.~\ref{eq:subslow} are:
$\vz^{(t)}, \vq^{(t)} = \SlowSubnet(\vx^{(t)}, \vy^{(t-1)}, \mW^{(t-1)}; \vtheta_{\text{slow}})$
with $\vz^{(t)}= (\vk^{(t)}, \vv^{(t)}, \beta^{(t)}, \bar{\vv}^{(t)})$.

\subsection{Related Models}

All the RFWP models presented in Sec.~\ref{sec:fast-net} and \ref{sec:slow-net}
can be seen as a type of memory augmented recurrent neural networks \citep{graves2014neural, graves2016hybrid}
in the sense that they maintain
two-dimensional fast weight states as a short-term memory, in addition to the standard one-dimensional RNN states.

There are also several previously proposed recurrent fast weight models.
For example, Schmidhuber's recurrent FWP from 1993 \cite{schmidhuber1993reducing}
has been revisited by \citet{ba2016using}.
There, key and value vectors are not generated within the same time step,
unlike in our models or in linear Transformers.
The Fast Weight Memory (FWM) \citep{schlag2020fastweightmemory}
is also a recurrent FWP: the slow net is an LSTM and the fast net is a higher-order RNN.
However, the FWM is a single pair of slow and fast nets, and a multi-layer version, as in the linear Transformer family, was not explored.
Similarly, the Metalearned Neural Memory \citep{MunkhdalaiSWT19} uses an LSTM as its slow net and a 3-layer MLP as its fast net but again limited to one pair.
Others have investigated variants of RNNs with fast weights
for toy synthetic retrieval tasks \citep{schlag2017gated, keller2018fast}.
In particular, \citet{keller2018fast} augment the LSTM with a fast weight matrix in the cell update.
In contrast, we make all weights in the LSTM fast and, importantly, our model specifications build upon the successful deep Transformer architecture using residual connections \citep{HeZRS16, srivastava2015icml}, layer-norm \citep{ba2016layer}, multiple attention heads and feed-forward blocks \citep{trafo}.
Essentially, we replace the self-attention
layers in the regular Transformers by
the fast weight programmer operations described above.

\section{Experiments}
\label{sec:exp}
We conduct experiments in four different settings.
We start by evaluating all models on a language modelling task (Sec.~\ref{sec:lm})
to obtain a performance overview and to discuss computational costs.
Language modelling is an excellent task to evaluate sequence models.
However, to highlight their different capabilities, we evaluate our models also on algorithmic tasks.
In fact, it is well-known that the actual capabilities of RNNs differ from one architecture to another \citep{weissGY18}.
We are interested in discussing such differences.
With that goal in mind, we conduct experiments on two synthetic algorithmic tasks,
code execution (Sec.~\ref{sec:lte}) and sequential ListOps (Sec.~\ref{sec:listops}),
which are designed to compare elementary sequence processing abilities of models.
Finally, we apply our models to reinforcement learning in 2D game environments (Sec.~\ref{sec:rl}) as a replacement for LSTMs.

\subsection{Language Modelling}
\label{sec:lm}
We first evaluate all discussed models on the generic language modelling task.
This allows for obtaining a performance overview and reviewing the computational efficiency of different models.
We use the Wikitext-103 dataset \citep{merity2016pointer} and follow the \textit{small model setting} similar to what's used in recent works by \citet{peng2021random} and \citet{schlag2021linear}.
This allows for training and evaluating different models with a reasonable amount of compute on this resource-demanding language modelling task.

\begin{table}[t]
\caption{WikiText-103 language model perplexity results with the \textit{small} setting \citep{peng2021random, schlag2021linear}.
For each model, its name, corresponding slow and fast networks, and weight update rule (Update) are specified.
All models are trained and evaluated on the span of 256 tokens
except for the models in the last two rows (+ full context) which
are trained and evaluated without context truncation. 
Parameter count is in millions.
See Appendix \ref{app:lm} for further experimental details and results.
}
\label{tab:lm_main}
\begin{center}
\begin{tabular}{llllrrr}
\toprule
Name & Slow net & Update & Fast net  & \multicolumn{1}{r}{Valid} & \multicolumn{1}{r}{Test} & \#Prms \\ \midrule
Transformer & - & -  &   -      &  33.0  & 34.1 & 44.0 \\ 
Linear Transformer &  Feedforward & sum  &   Linear      &  37.1  & 38.3  & 44.0 \\ 
Delta Net &  & delta &       & 34.1   & 35.2 & 44.0 \\ \midrule 
Delta MLP & Feedforward & delta &   Deep MLP    &  35.8  & 36.8 &  44.3 \\
Delta Delta Net & &  &   Delta Net    & 34.0   & 35.2 & 44.6 \\
Delta RNN &  &  &   RNN    &  33.8  &  35.0  & 44.6 \\
Delta LSTM &  &   &   LSTM    &  \textbf{32.6}  & \textbf{33.8}  & 47.3 \\
RDN & Recurrent &  &  Linear    & 34.1  & 35.2 &  44.1 \\ \midrule
Delta RNN &  \multirow{2}{*}{+ full context}  &   &  & \textbf{31.8}  & \textbf{32.8} &  44.6  \\ 
RDN & &   &  & 32.5  & 33.6 &  44.1 \\
\bottomrule
\end{tabular}
\end{center}
\end{table}

\paragraph{Perplexity results.}
The results are shown in Table \ref{tab:lm_main} which
also serves as a tabular summary 
recapitulating different models described in Sec.~\ref{sec:background} and \ref{sec:model},
with various architectures for slow and fast nets, and two choices of update rule.
The top block of Table \ref{tab:lm_main} shows the performance of the baseline
Transformer, \citet{katharopoulos2020transformers}'s Linear Transformer,
and \citet{schlag2021linear}'s Delta Net.
The performance of models presented in Sec.~\ref{sec:model}
can be found in the middle block.
First of all, the Delta MLP
performs worse than the baseline Delta Net despite a slight increase in parameter count (44.3 vs.~44.0\,M).
This supports the intuition that it is better to make the slow network aware of
the outputs of intermediate layers to generate fast weights in a deep network,
instead of generating fast weights for all layers at a time.
In all other models, the performance never degrades
with the proposed architectural augmentation.
The Delta Delta Net yields limited improvements;
we plan to study this model in depth in future work.
With the same amount of parameters (44.6~M),
the Delta RNN yields greater improvements.
Among the models presented here, the Delta LSTM variant exhibits the best
performance.
This shows that the slow network successfully controls the rather complex fast LSTM network, although it also requires more parameters (47.3~M)
than other models.
Finally, the benefits of recurrent connections added to the baseline Delta Net
do not directly translate into practical improvements in language modelling as
demonstrated by the performance of RDN compared to the one of the baseline Delta Net.
Importantly, given a constant memory size w.r.t.~sequence length,
it is straight-forward to train and evaluate our RNNs without context truncation
(while still limiting the backpropagation span).
Corresponding performances of Delta RNN and RDN
are shown in the bottom part of Table \ref{tab:lm_main}: they outperform the regular Transformer with a limited context (256 tokens).

While language modelling is useful as a sanity check 
(here for example, except for the Delta MLP, all models
achieve reasonable performance),
the task is too generic to identify certain important aspects of the models,
such as real benefits of recurrence.
Before we move on to trickier RL applications, Sec.~\ref{sec:lte} and \ref{sec:listops} will focus on studying such aspects using synthetic algorithmic tasks.

\paragraph{Computational efficiency.}
The modifications we proposed in Sec.~\ref{sec:model}
introduce additional computational costs
to linear Transformers/FWPs.
First of all, none of them affect the core complexity of linear Transformers:
they all have a constant space and linear time complexity
w.r.t.~sequence length.
However, the per-time-step computational costs differ a lot from one
model to another, as quantified here in terms of 
training speed using our implementation.
All models are implemented using a custom CUDA kernel except the baseline
Transformer for which we use regular PyTorch code \citep{NEURIPS2019_bdbca288}.
Training speeds of LT and Delta Net in Table \ref{tab:lm_main}
are 66\,K and 63\,K words per second respectively
(vs.~33\,K for the baseline Transformer).
The most expensive model is the Delta LSTM.
This fast weight LSTM with tied input-forget gates has 6 weight matrices,
and each of these are manipulated by separate delta rules.
The corresponding speed is 14\,K words per second,
too slow for scaling to more experiments.
In contrast, the speeds of Delta RNN and RDN remain
reasonable:  41\,K and 35\,K words per second respectively.
Therefore, the remaining experiments will focus on these two recurrent architectures which are
promising and practical in terms  of both performance and computational
costs.

\subsection{Code Execution Task: Learning to Maintain and Update Variable States}
\label{sec:lte} 
In code execution tasks \citep{zaremba2014learning}, models are trained to sequentially read the
input code provided as word-level text, and to predict the results of the corresponding code execution.
We adopt the task setting from \citet{fan2020addressing} with one conditional and three basic statements.
We refer the readers to Appendix \ref{app:lte_task} for a precise description of the task.
This code execution task requires models to maintain the values of multiple variables,
which has been shown to be difficult for relatively shallow Transformers
with only feedforward connections \citep{fan2020addressing}.

The left block of Table \ref{tab:exp_algo} shows the results.
Following again \citet{fan2020addressing}, we control the task difficulty
by modifying the number of variables (3 or 5).
The model architectures are fixed: the LSTM has only one layer with 256 nodes
and all Transformer variants have the same architecture with 4 layers
with a hidden size of 256 using 16 heads and an inner feedforward layer size of 1024.

We first note that the LSTM is the best performer for both difficulty levels, with the smallest performance drops through increasing the number of variables.
In contrast to prior claims \citep{fan2020addressing}, the LSTM is clearly capable of storing the values of multiple variables in its hidden and cell state vectors.
With three variables, the regular Transformer already largely underperforms
other models with a mutable memory: Delta Net, Delta RNN, and RDN.
Linear Transformers completely fail at this task, likely due to the memory capacity problem pointed out by \citet{schlag2021linear} (see Appendix \ref{app:lte_exp}
for further discussion).
By increasing the number of variables to five,
the baseline Transformers, Delta Net, and RDN become unstable
as shown by high standard deviations w.r.t. the seed.
The benefits of recurrent connections introduced in our RDN
compared to the baseline Delta Net become more apparent (76.3 vs.~61.4\%).
In contrast, the Delta RNN remains stable and gives the best performance (85.1\%)
among Transformer variants, which shows the benefits of recurrence and in particular the regular RNN architecture in the fast net.
To match the performance of LSTM on this task, however,
these models need more layers  (see Appendix \ref{app:lte_exp} for more results).

\begin{table}[h]
\caption{\textbf{Test accuracies} (\%) with standard deviations on \textbf{code execution} (Code Exec) and \textbf{sequential ListOps} (Seq ListOps).
The difficulty of the task is controlled by the maximum number 
of possible variables (\# variables) for code execution,
and the list depth (10 or 15) for ListOps.
For code execution with 5 variables, we report means over six seeds.
In all other cases, the results are computed with three seeds.
For more results, see Appendix \ref{app:lte_exp} (Code Exec) and \ref{app:listops} (Seq ListOps).}
\label{tab:exp_algo}
\begin{center}
\begin{tabular}{lrrrr}
\toprule
&    \multicolumn{2}{c}{\textbf{Code Exec} (\# variables)} & \multicolumn{2}{c}{\textbf{Seq ListOps} (depth)} \\ \cmidrule(r){2-3}   \cmidrule(r){4-5}  
&    \multicolumn{1}{c}{3} & \multicolumn{1}{c}{5}   & \multicolumn{1}{c}{10} & \multicolumn{1}{c}{15} \\
\midrule
LSTM        &     \textbf{99.0} $\pm$ {\small 0.1}  & \textbf{93.2}  $\pm \,\;$ {\small6.1}  & \textbf{88.5} $\pm$ {\small 2.9}  & 24.4 $\pm$ {\small 1.1} \\
Transformer            &    71.8  $\pm$ {\small 2.6} &   35.4  $\pm$ {\small 28.2} &  79.1  $\pm$ {\small 0.9}    & 75.3  $\pm$ {\small 0.4}  \\
Linear Transformer      &   0.0  $\pm$ {\small 0.0}   & 0.0  $\pm \,\;$ {\small 0.0}  & 64.0  $\pm$ {\small 0.3}    & 64.4  $\pm$ {\small 0.4}  \\
Delta Net            &  90.7  $\pm$ {\small 2.7} &  61.4 $\pm$ {\small 20.0} &  \textbf{85.7}  $\pm$ {\small 1.8}  &  77.6   $\pm$ {\small 1.4}    \\ \midrule 
Delta RNN    &   90.8  $\pm$ {\small 1.7}   & \textbf{85.1}  $\pm \,\;$ {\small 1.9} &    83.6  $\pm$ {\small 1.2}  &  78.0  $\pm$ {\small 1.0} \\
RDN  &   \textbf{92.6}  $\pm$ {\small 2.2} &   76.3  $\pm$ {\small 17.6}   &   83.2  $\pm$ {\small 0.9}   &  \textbf{79.2}  $\pm$ {\small 1.4} \\
\bottomrule
\end{tabular}
\end{center}
\end{table}

\subsection{Sequential ListOps: Learning Hierarchical Structure and Computation}
\label{sec:listops}
The ListOps task \citep{nangia-bowman-2018-listops}
is a typical test for hierarchical structure learning, which requires list operation executions.
We use a simple variant of ListOps whose
detailed descriptions can be found in Appendix \ref{app:listops}.
For example, the list \texttt{[MAX 6 1 [FIRST 2 3 ] 0 [MIN 4 7 1] ]} is of depth two and the expected output is \texttt{6}.
While early research comparing self-attention to RNNs \citep{TranBM18} has shown
some advantages of recurrence in hierarchical structure learning,
more recent work \citep{lu2021pretrained}
reports Transformers outperforming LSTMs on ListOps. According to 
\citet{tay2020long}, linear Transformer variants (LT and Performers) underperform other Transformer variants by a large margin on ListOps.

The right block of Table \ref{tab:exp_algo} shows results for two different depths:~10 and 15.
The model architectures are identical to those used in the code execution task (Sec.~\ref{sec:lte}).
At depth 10, we find LSTM to perform best, while mutable memory Transformer variants
(Delta Net, Delta RNN, and RDN)
outperform the regular and linear Transformers.
At depth 15, the LSTM's performance drops drastically (to 24.4\%),
while the differences between Transformer variants remain almost the same.
We note that sequences are longer
for the depth 15 problem (mean length of 185 tokens) than for the depth 10 version (mean length of 98 tokens).
This turns out to be difficult for the small 256-dimensional LSTM;
see Appendix \ref{app:listops}
for the corresponding ablation study.
The performance differences between the baseline Delta Net and the proposed Delta RNN and RDN are rather small for this task.
Importantly, 
our models outperform both regular and linear Transformers
on this task requiring hierarchical structure learning.

\subsection{Reinforcement Learning in 2D Game Environments}
\label{sec:rl}
We finally evaluate the performance of our models as a direct replacement for the LSTM in reinforcement learning settings.
In fact, only a limited number of prior works have investigated 
Transformers for RL.
\citet{parisotto2020stabilizing} and \citet{rae2020compressive} evaluate them on the DMLab-30 \citep{beattie2016deepmind, leibo2018psychlab}.
\citet{parisotto2020stabilizing} also evaluate them on Atari but in a multi-task setting \citep{BadiaPKSVGB20}.
Others \citep{fan2020addressing, parisotto2021efficient} use toy maze environments.
In contrast to \citet{parisotto2020stabilizing}'s work, which presents multi-task Atari
as a side experiment, we study the Transformer family of models on the standard Atari 2600 setting \citep{BellemareOGTM16, mnih2013playing, mnih2015human}
by training game-specific agents.
\begin{figure}[h]
    \begin{minipage}{0.45\textwidth}
    \begin{center}
        \includegraphics[width=1.05\columnwidth]{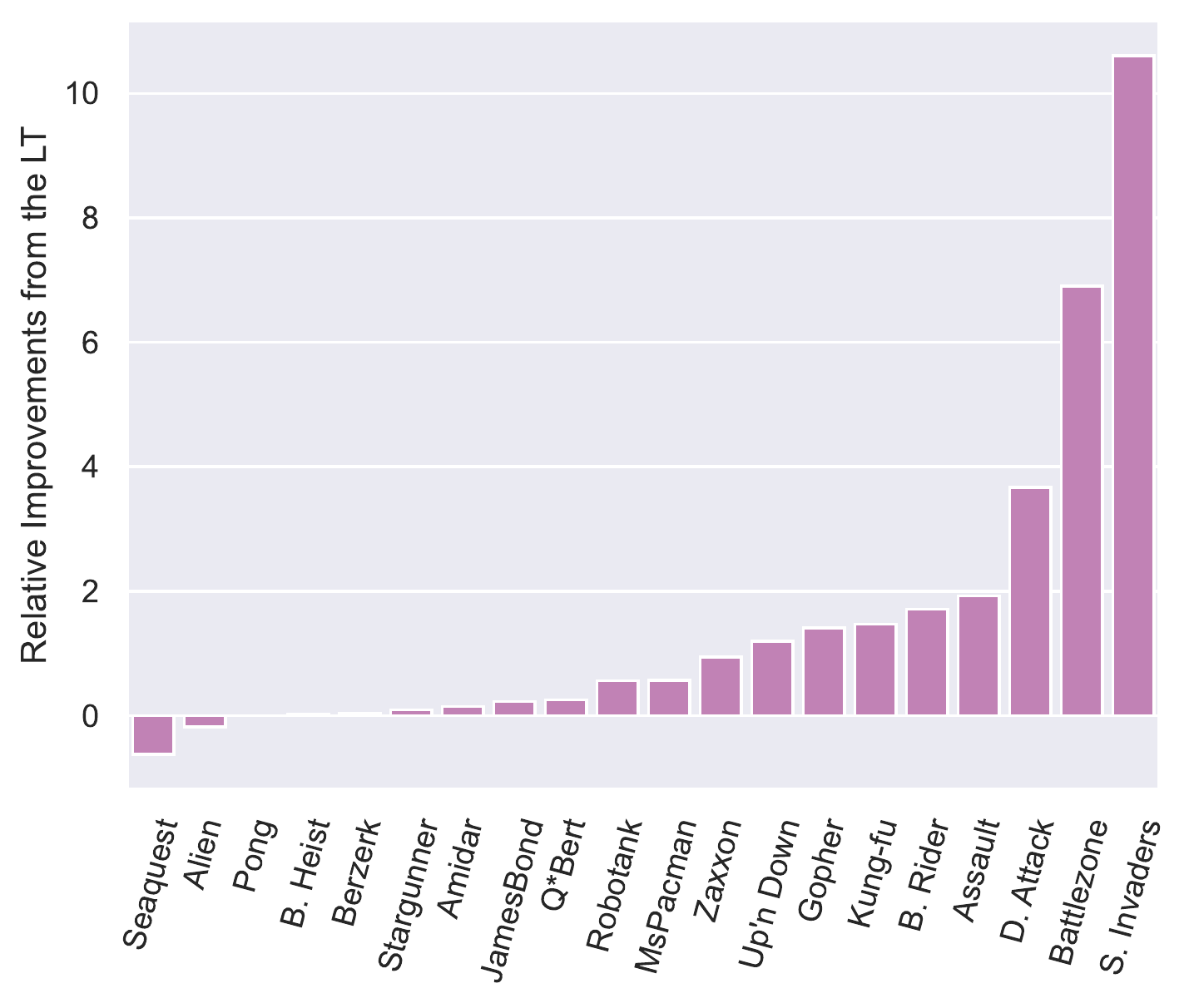}
        \caption{Relative improvements in test scores obtained by the \textbf{Recurrent Delta Net (RDN)} compared to the \textbf{Linear Transformer (LT)} after \textbf{50\,M} env.~steps.}
        \label{fig:atari_lt_50m}
    \end{center}
\end{minipage}
\hspace{8mm}
    \begin{minipage}{0.45\textwidth}
    \begin{center}
        \includegraphics[width=1.05\columnwidth]{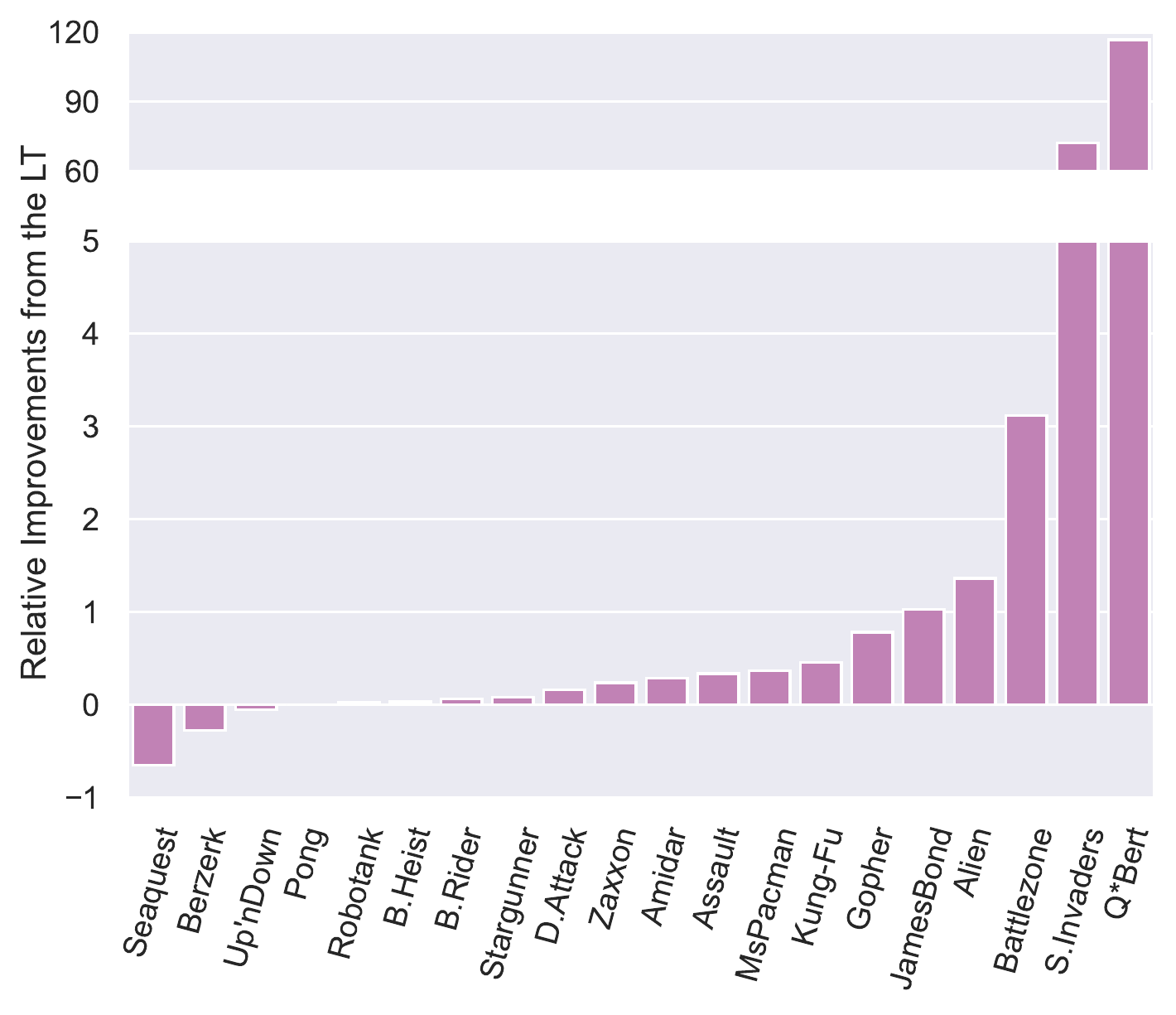}
        \caption{Relative improvements in test scores obtained by the \textbf{Recurrent Delta Net (RDN)} compared to the \textbf{Linear Transformer (LT)} after \textbf{200\,M} env.~steps.}
        \label{fig:atari_lt_200m}
    \end{center}
\end{minipage}
\end{figure}
%
\begin{figure}[b]
    \begin{minipage}{0.45\textwidth}
    \begin{center}
        \includegraphics[width=1.05\columnwidth]{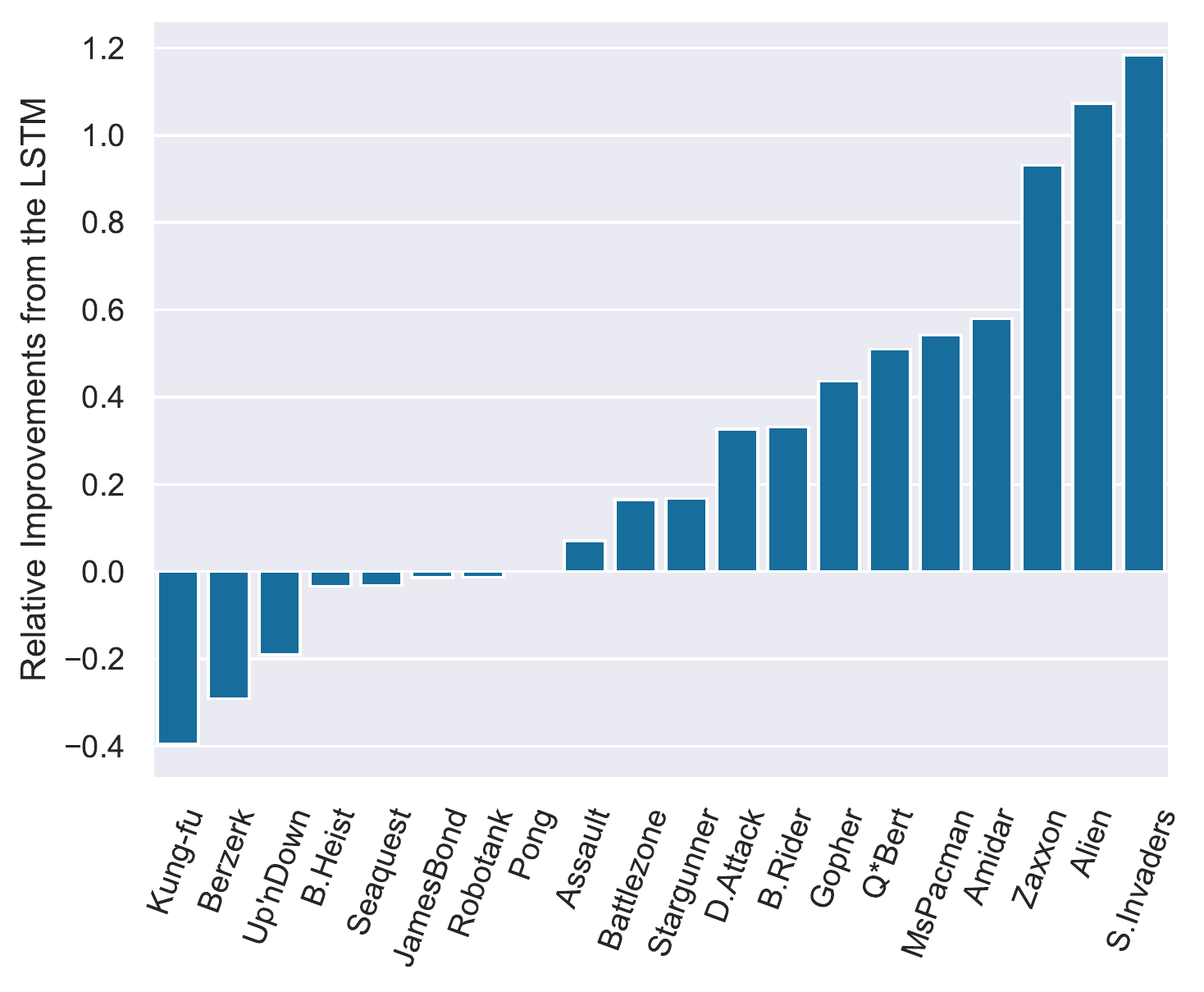}
        \caption{Relative improvements in test scores obtained by 2-layer \textbf{RDN} compared to \textbf{LSTM} after \textbf{50\,M} env.~steps.}
        \label{fig:atari_50m}
    \end{center}
\end{minipage}
\hspace{8mm}
    \begin{minipage}{0.45\textwidth}
    \begin{center}
        \includegraphics[width=1.05\columnwidth]{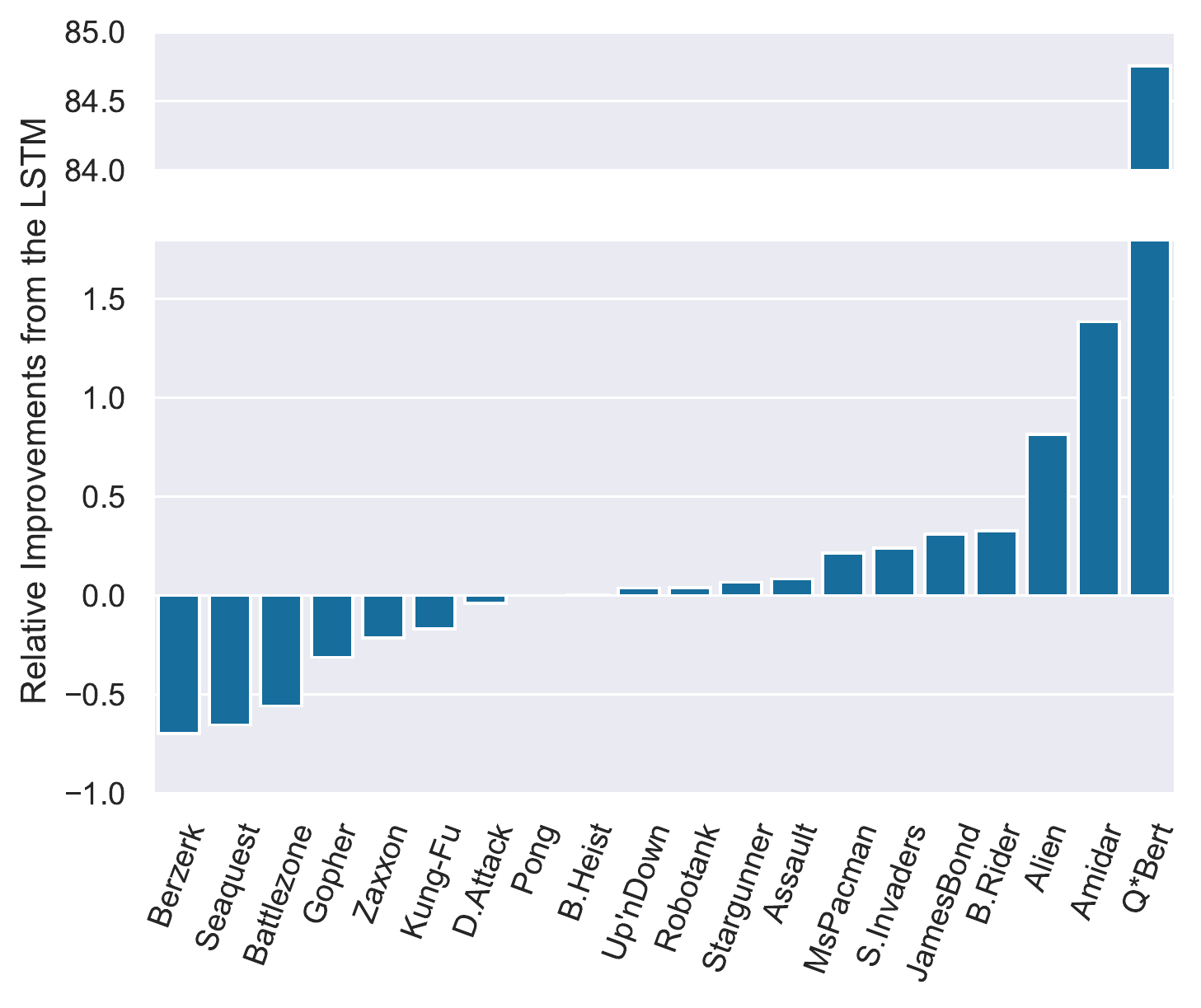}
        \caption{Relative improvements in test scores obtained by 2-layer \textbf{RDN} compared to \textbf{LSTM} after \textbf{200\,M} env.~steps.}
        \label{fig:atari_200m}
    \end{center}
\end{minipage}
\end{figure}
\paragraph{Settings.} 
We train an expert agent on each game separately with the Importance Weighted Actor-Learner Training Architecture (IMPALA) using the V-trace actor-critic setup \citep{EspeholtSMSMWDF18} and entropy regularization \citep{MnihBMGLHSK16} implemented in \texttt{Torchbeast} \citep{kuttler2019torchbeast}.
Our model follows the \textit{large} architecture of \citet{EspeholtSMSMWDF18} which consists of a 15-layer residual convolutional NN with one 256-node LSTM layer which we replace by either the RDN (Sec.~\ref{sec:slow-net}) or the Delta RNN (Sec.~\ref{sec:fast-net}).
In line with the small LSTM used for Atari (only 1 layer with 256 hidden nodes)
we also configure a small RDN:
2 layers with a hidden size of 128 using 4 heads, and a feedforward dimension of 512.
We find this small model to perform already surprisingly well.
For the rest, we use the same hyperparameters as \citet{EspeholtSMSMWDF18}
which can be found in Appendix \ref{app:rl}.

\paragraph{Main experiments.} 
We evaluate our models in 20 environments. 
According to \citet{MottZCWR19}, in about half of them, the LSTM outperforms the feedforward baselines---which we confirm in our setting with 50\,M steps (see Appendix \ref{app:rl}). 
We report results at 50\,M and 200\,M environmental steps of training.
Like \citet{nair2015massively}, we run the trained agent for 30 test episodes.
Here we repeat this evaluation five times to report the average score with a standard deviation.
The following analysis focuses on the RDN (Sec.~\ref{sec:slow-net}) compared to
the regular linear Transformer and the LSTM.
A similar study of the Delta RNN,
as well as comparisons to more baselines,
and the exact scores achieved by each model
on each game can be found in Appendix \ref{app:rl}.

In all our experiments above, we have shown that the Linear Transformer,
i.e., a Fast Weight Programmer with a purely additive update rule,
consistently underperforms other models based on the delta rule.
Here we confirm this trend once more.
Figures \ref{fig:atari_lt_50m} and \ref{fig:atari_lt_200m} show
the relative improvements of scores obtained by Recurrent Delta Net over those achieved by the linear Transformer on each game, respectively after 50 and 200\,M interaction steps.
The RDN matches or outperforms the Linear Transformer on all games except for two out of 20 games at both stages of training.

Figure \ref{fig:atari_50m} shows relative improvements of RDN over LSTM after 50\,M interactions.
In 12 games, the RDN yields improvements over LSTM, whereas in 3 games, the LSTM performs better.
In the remaining 5 games, both reach similar scores.
Interestingly, this trend does not directly extrapolate to the 200\,M case, which is presented in Figure \ref{fig:atari_200m}.
With longer training, the LSTM surpasses the performance of the RDN in \textit{Battlezone}, \textit{Gopher}, \textit{Seaquest} and \textit{Zaxxon},
while the RDN catches up in \textit{Up'N Down} and \textit{Kung-Fu Master}.
Overall, there are 6 games in which LSTM clearly outperforms RDN at 200\,M steps,
whereas in 9 games the result is the opposite.

On a side note, some of the scores achieved by the RDN at 200\,M step are excellent:
a score of over 170\,K and 980\,K in \textit{Space Invader} and \textit{Q*Bert} respectively beats the state-of-the-art set by MuZero \citep{schrittwieser2020mastering} and Agent57 \citep{BadiaPKSVGB20}.
However, a direct comparison is not fair as we train game-specific agents.

\paragraph{Experiments with larger models.}
Given the results above, a natural question to ask is whether a larger model size improves the RDN in games where the LSTM dominates.
We focus on four such games: \textit{Battlezone}, \textit{Berzerk}, \textit{Gopher}, and \textit{Seaquest} (See Fig.~\ref{fig:atari_200m}).
We double the model size to 3.4\,M parameters by increasing the number of layers to 4 and the hidden size to 256, with 8 heads.
As shown in Table \ref{tab:atari_deeper}, larger RDN models reduce the gap to the LSTM
(except in \textit{Berzerk}).
This indicates that further scaling RDN might be as promising as scaling regular Transformers in other domains.

\begin{table}[h]
\caption{Performance of a larger RDN in \textbf{games where the LSTM dominates} (200\,M steps).
}
\label{tab:atari_deeper}
\begin{center}
\begin{small}
\begin{tabular}{lrrrr}
\toprule
   & \multicolumn{1}{c}{Battlezone} &  \multicolumn{1}{c}{Berzerk}  &  \multicolumn{1}{c}{Gopher}  & \multicolumn{1}{c}{Seaquest}  \\ \midrule
LSTM        &  \mpm{24,873}{1,240}    &  \mpm{\textbf{1,150}}{92}   & \mpm{\textbf{124,914}}{22,422}  &  \mpm{12,643}{1,627} \\
RDN      &  \mpm{10,980}{1,104}   &  \mpm{348}{17}   & \mpm{86,008}{11,815}  & 4,373 $\pm \,\;\;$ 504 \\
RDN larger    &  \mpm{\textbf{28,273}}{5,333}    & 346 $\pm \,\;$ 9   &  \mpm{118,273}{14,872}  & \textbf{14,601}  $\pm \,\;\;$ {\small 712}  \\
\bottomrule
\end{tabular}
 \end{small}
\end{center}








\end{table}

\section{Conclusion}
Inspired by the formal equivalence of linear Transformers and certain traditional Fast Weight Programmers (FWPs) from the early '90s, we propose various new linear Transformer variants with recurrent connections.
Our novel Recurrent FWPs (RFWPs) outperform previous linear and regular Transformers on a code execution task and significantly improve over Transformers in a sequential ListOps task.
On Wikitext-103 in the ``small'' model setting, RFWPs compete well with the previous best linear Transformer variants for truncated contexts, and with full contexts, beat regular Transformers.
Our RFWPs can also be used as drop-in replacements for problems where RNNs are still dominant. 
In particular, we evaluate them in reinforcement learning settings on 20 Atari 2600 environments.
They clearly outperform the regular Linear Transformer on almost all environments.
They also outperform the LSTM across many environments with a small model size and demonstrate promising scaling properties for larger models.
Given the increasing interest in deploying Transformers in RL \citep{chen2021decision, janner2021reinforcement}, in particular in the framework
of Upside-Down RL \citep{schmidhuber2019reinforcement, srivastava2019training}, our RFWP models are particularly relevant: as RNNs, they conveniently handle long contexts with a constant memory size, while being powerful Transformer variants at the same time.
Our work highlights the usefulness of the FWP framework from the '90s and its connection to modern architectures, opening promising avenues for further research into new classes of recurrent Transformers. 

\begin{ack}
We thank Aleksandar Stani\'c and Sjoerd van Steenkiste for valuable comments on the first version of this paper.
This research was partially funded by ERC Advanced grant no: 742870, project AlgoRNN,
and by Swiss National Science Foundation grant no: 200021\_192356, project NEUSYM.
This work was partially supported by
computational resources at the CSCS Swiss
National Supercomputing Centre, project d115.
We thank NVIDIA Corporation for donating several DGX
machines, and IBM for donating a Minsky machine.
\end{ack}

\bibliography{references}

\bibliographystyle{unsrtnat}

\clearpage
\appendix

\section{Experimental Details and Ablation Studies for Language Modelling}
\label{app:lm}

\subsection{Experimental Settings}
All language models in Table \ref{tab:lm_main} have the same Transformer configuration:
a 16-layer model with
a hidden size of 128 with 8 heads, and a feed-forward dimension of 2048.
We use a dropout \citep{hanson1990stochastic, srivastava2014dropout, frazier2018dropout} rate of 0.1.
The batch size is 96 and we train for
about 120 epochs with Adam optimiser \citep{kingma2014adam} with an initial
learning rate of 0.00025 and 2000 learning rate warm-up steps. 
All models are trained with a back-propagation span of 256 tokens.
During training, these segments are treated independently, except for the
\textit{+ full context} cases in Table \ref{tab:lm_main}
where the states (both recurrent states and fast weight states) 
from a segment are used as initialisation for the subsequent segment.
The models in \textit{+ full context} cases are also evaluated in the same way
by carrying over the context throughout the evaluation text with a batch size of one.
For all other cases, the evaluation is done by going through the text
with a sliding window of size 256 with a batch size of one.
Transformer states are computed for all positions in each window,
but only the last position is used to compute perplexity (except in the first
segment where all positions are used for evaluation) \citep{al2018character}.
We trained all models using two GPUs (32~GB V100),
and the longest training takes up to 10 days
(see Sec.~\ref{sec:lm} in the main text for speed comparison between models).

For readers interested in any further details, we refer to our code
which is publicly available.

\subsection{Ablation Studies}
\label{app:lm_models}
In this section, we specify the exact Delta LSTM and Delta MLP models used in Table \ref{tab:lm_main},
and provide a few ablation studies
for Delta RNN, Delta LSTM\footnote{The numbers reported in Table \ref{tab:lm_ablation} for the Delta LSTM models are better than those we presented in an earlier version.
In fact, we found that in our previous code,
the slow weights for key and value generation 
were shared by mistake between the forward and recurrent fast weight matrices (while the reported parameter count was that of the correct model with separate slow weight matrices).
Fixing this resulted in the corresponding improvements.}, and Delta MLP models.

\begin{table}[h]
\caption{Ablation studies for Delta LSTM, Delta RNN, and Delta MLP models.
Language model perplexities are shown and the setting is the same as in Table \ref{tab:lm_main}.
}
\label{tab:lm_ablation}
\begin{center}
\begin{tabular}{lcrrr}
\toprule
 & Version & \multicolumn{1}{r}{Valid} & \multicolumn{1}{r}{Test} & \#Prms \\ \midrule
Delta RNN &  A    &  35.6 & 36.7 & 44.6  \\ 
 &  B    &  \textbf{33.8}  &  \textbf{35.0}  & \\  \midrule
Delta LSTM &  A    & 38.5  & 39.9 &  47.3 \\
&  B    & 34.2  & 35.2 &  \\
&  C    & 33.5 & 34.7  &  \\
&  D    &  \textbf{32.6}  & \textbf{33.8} & \\  \midrule
Delta MLP &  A   &36.8  &  37.9 & 44.3 \\   
 &  B    &  \textbf{35.8} &  \textbf{36.8} & 44.3   \\   
\bottomrule
\end{tabular}
\end{center}

\end{table}

\paragraph{Delta RNN.}
In Sec.~\ref{sec:fast-net}, we argue for a version of fast RNN given by Eq.~\ref{eq:fastrnn}
as a natural augmentation of the linear Transformer with recurrent connections.
Here we empirically support this choice by comparing to another variant of Delta RNN given by:
\begin{eqnarray}
\label{eq:fastrnn_v2}
\vy^{(t)}  &=& f(\mW^{(t)} \vq^{(t)} + \mR^{(t)} \vy^{(t-1)})
\end{eqnarray}
where  $f$ is again the softmax activation which makes $\vy^{(t)}$
a valid query vector (positive components which sum up to one)
for fast weights maintained by the delta update rule.
We refer to this version as \textbf{Version A}
in this ablation study and the one given by Eq.~\ref{eq:fastrnn} as \textbf{Version B}.
As Table \ref{tab:lm_ablation} shows, Version B performs 
better, and this is the one we report in Table \ref{tab:lm_main} in the main text. 
 
\paragraph{Delta LSTM.}
We evaluate four versions of Delta LSTM for ablation.
In all cases, we tie the input and forget gates to
reduce the total number of fast weights to be controlled by the slow net.
All models contain six fast weights and each of them is updated according 
to the delta update rule (Eq.~\ref{eq:delta}).
The different versions differ in the location of
the activation function and residual connections
in the LSTM architecture \citep{hochreiter1997long, gers2000learning},
inspired by the Transformer.

\textbf{Version A} (analogous to Version A of Delta RNN above)
is the one which is the closest to the original LSTM
with tied input and forget gate.
The only architectural difference is the usual $\tanh$ on the cell output $\vc^{(t)}$ which is replaced by 
a softmax $f$ placed after the final output of the layer $f(\vc^{(t)} \odot \vo^{(t)})$,
such that it can directly be used as a delta rule compatible query for the next time step
(we also use a sigmoid instead of $\tanh$ for the main transformation $\vu^{(t)}$,
but this is not crucial for any models here).
\begin{eqnarray}
\label{eq:lstm_v1}
\vu^{(t)} &=&  \sigma(\mW^{(t)} \vq^{(t)} + \mR^{(t)} \vy^{(t-1)}) \\
\vf^{(t)} &=&  \sigma(\mW_f^{(t)} \vq^{(t)} + \mR_f^{(t)} \vy^{(t-1)}) \\
\vo^{(t)} &=&  \sigma(\mW_o^{(t)} \vq^{(t)} + \mR_o^{(t)} \vy^{(t-1)}) \\
\vc^{(t)} &=& \vf^{(t)} \odot \vc^{(t-1)} +  (1 - \vf^{(t)}) \odot \vu^{(t)} \\
\vy^{(t)} &=& f(\vc^{(t)} \odot \vo^{(t)})
\end{eqnarray}

\textbf{Version B} is obtained by delaying the application of the
softmax activation $f$ in Version A.
\begin{eqnarray}
\label{eq:lstm_v2}
\vu^{(t)} &=&   \sigma(\mW^{(t)} \vq^{(t)} + \mR^{(t)} f(\vy^{(t-1)})) \\
\vf^{(t)} &=&  \sigma(\mW_f^{(t)} \vq^{(t)} + \mR_f^{(t)} f(\vy^{(t-1)})) \\
\vo^{(t)} &=&  \sigma(\mW_o^{(t)} \vq^{(t)} + \mR_o^{(t)} f(\vy^{(t-1)})) \\
\vc^{(t)} &=& \vf^{(t)} \odot \vc^{(t-1)} +  (1 - \vf^{(t)}) \odot \vu^{(t)} \\
\vy^{(t)} &=& \vc^{(t)} \odot \vo^{(t)} 
\end{eqnarray}

\textbf{Version C} is obtained from Version B by adding a residual connection from the feed-forward
part $\vz_u^{(t)}$ of the main transformation $\vu^{(t)}$ to the output.
\begin{eqnarray}
\label{eq:lstm_v3}
\vz_u^{(t)} &=& \mW^{(t)} \vq^{(t)} \\
\vu^{(t)} &=&   \sigma(\vz_u^{(t)} + \mR^{(t)} f(\vy^{(t-1)})) \\
\vf^{(t)} &=&  \sigma(\mW_f^{(t)} \vq^{(t)} + \mR_f^{(t)} f(\vy^{(t-1)})) \\
\vo^{(t)} &=&  \sigma(\mW_o^{(t)} \vq^{(t)} + \mR_o^{(t)} f(\vy^{(t-1)})) \\
\vc^{(t)} &=& \vf^{(t)} \odot \vc^{(t-1)} +  (1 - \vf^{(t)}) \odot \vu^{(t)} \\
\vy^{(t)} &=& \vc^{(t)} \odot \vo^{(t)}  + \vz_u^{(t)}
\end{eqnarray}

Finally, 
\textbf{Version D} is obtained from Version B by removing the sigmoid on the main transformation
$\vu^{(t)}$ which results in a highway net-like skip connection \citep{srivastava2015icml} from $\vu^{(t)}$ to the output.
This version is then analogous to Version B of the Delta RNN as a natural augmentation of the
linear Transformer: a recurrent term is added to the main transformation $\vu^{(t)}$
and gating components are added to make it an LSTM architecture:
\begin{eqnarray}
\label{eq:lstm_v4}
\vu^{(t)} &=&  \mW^{(t)} \vq^{(t)} + \mR^{(t)} f(\vy^{(t-1)}) \\
\vf^{(t)} &=&  \sigma(\mW_f^{(t)} \vq^{(t)} + \mR_f^{(t)} f(\vy^{(t-1)})) \\
\vo^{(t)} &=&  \sigma(\mW_o^{(t)} \vq^{(t)} + \mR_o^{(t)} f(\vy^{(t-1)})) \\
\vc^{(t)} &=& \vf^{(t)} \odot \vc^{(t-1)} +  (1 - \vf^{(t)}) \odot \vu^{(t)} \\
\vy^{(t)} &=& \vc^{(t)} \odot \vo^{(t)} 
\end{eqnarray}

Corresponding performances can be found in Table \ref{tab:lm_ablation}.
The best model, Version D, is the one we report in Table \ref{tab:lm_main} in the main text.

\paragraph{Delta MLP.}
We also conduct a few ablation studies for the Delta MLP (Sec.~\ref{sec:fast-net}).
As MLP architecture we used the feedforward block of the regular Transformer
which consists of two feedforward layers: one with the size of the inner feedforward layer (2048 here) and another one with the size of hidden dimension (128 here).
We test two variants which result in a similar number of parameters:
\textbf{Version A} with 8 overall Transformer layers where each self-attention layer contains 4 fast MLP layers (i.e.~a total of 48 feedforward layers with 32 fast ones), and \textbf{Version B} with 11 overall Transformer layers where each self-attention layer contains 2 fast MLP layers (i.e.~a total of 44 feedforward layers with 22 fast ones).
As shown in Table \ref{tab:lm_ablation}, Version B which has fewer fast layers controlled by the same slow net performs better,
and, as already mentioned in Sec.~\ref{sec:exp}, they do not outperform the baseline Delta Net which has only one fast feedforward layer (Table \ref{tab:lm_main}).

\subsection{Dimensionality of Delta-Delta Net vs. Delta Net}
\label{app:deltadelta}
Here we describe how the dimensionality of
Delta-Delta Net scales with the size of the Delta Net.
We assume a Delta Net with a dimension $d$ for all query, key, value and input vectors.
Then its slow weight matrix (the projection matrix) is of size $d \times (3d+1)$ as it projects a $d$-dimensional input to query, key, value vectors ($3d$) and a scalar beta ($+1$) which are needed to maintain a $d \times d$ fast weight matrix using the delta rule.
Now we can express the dimensionality of a Delta-Delta Net in terms of $d$, whose fast network is a Delta Net with the dimensionality above.
The size of its fast weight matrix is thus $d \times (3d+1)$.
In order to maintain a fast weight matrix of this dimension using the delta rule, we need key and query vectors of size $d$, a value vector of size $3d+1$, and a scalar beta ($+1$).
The slow weight matrix has to produce all these variables with a total dimension of $(5d+2)$ from the input of size $d$.
Therefore, the size of the slow weight matrix in the Delta-Delta Net is $d \times (5d+2)$.
Such a Delta-Delta Net would have to store two fast weight matrices: one of size $d \times (3d+1)$ and another one of size $d \times d$.

\section{Experimental Details and Additional Results for Algorithmic Tasks}
\label{app:algo}

\subsection{Task Details for Code Execution}
\label{app:lte_task}
In code execution tasks \citep{zaremba2014learning}, models are trained to sequentially read the
input code provided as word-level text and to predict the results of the corresponding code execution.
We adopt the task setting from \citet{fan2020addressing}.
Each example is a sequence consisting of multiple statements --- 100 in our experiments.
A statement can be one of the following three basic statements:
\texttt{assign} which assigns a value to a variable (e.g. \texttt{x = 2 ;}),
\texttt{increment} which increments or decrements an already assigned variable
(e.g. \texttt{x ++ ;}),
or \texttt{print} which outputs the value of the variable (e.g. \texttt{print x ;}).
In addition to basic statements, there are also conditional comparisons on already defined variables followed by a basic statement
(e.g. \texttt{if x < 3 :~x ++ ;}).
The model reads the input word-level code sequence from left to right
in an auto-regressive manner, and makes a prediction at each position: at the end of each \texttt{print}
statement, the model has to predict the correct variable value,
and for all other positions, the no-output token.

Here is a short example (with \texttt{N} denoting the no-output token):

\begin{table}[h]
\small		
\label{tab:example_lte}
\begin{center}
\setlength{\tabcolsep}{0.5em}
\begin{tabular}{lcccccccccccccccccccccc}
In: & \texttt{x} & \texttt{=} & \texttt{3} & \texttt{;} & \texttt{y} & \texttt{=} & \texttt{7} & ; & x &++ &  \texttt{;} & \texttt{if} & \texttt{y} & \texttt{<} & \texttt{6} & \texttt{:} & \texttt{print} & \texttt{x} & \texttt{;} & \texttt{print} & \texttt{x} & \texttt{;} \\
Out: & \texttt{N} & \texttt{N} & \texttt{N} & \texttt{N} & \texttt{N} & \texttt{N} & \texttt{N} & \texttt{N} & \texttt{N} & \texttt{N} & \texttt{N} & \texttt{N} & \texttt{N} & \texttt{N} & \texttt{N} & \texttt{N} & \texttt{N} & \texttt{N} & \texttt{N} & \texttt{N} & \texttt{N} & \texttt{4} \\
\end{tabular}
\end{center}
\end{table}
In contrast to \citet{fan2020addressing}, we hard-code the last statement
to be a \texttt{print} statement of a randomly chosen variable such that the model always has to make a prediction at the end of the sequence.
The output vocabulary of the model is restricted to discrete values within a
pre-determined range (here between -8 and 16), and
the code sequences are constructed such that the value to be printed does not
exceed this range by rejecting any statement which would result in such values.
Like \citet{fan2020addressing}, we randomly generate 10,000 sequences for training and 1,000 sequences each for validation and test.
With 100 statements per sequence, we obtain sequences with lengths varying from about 370 to 550,
with an average length of about 450 tokens for both 3 and 5 variable cases, and for train, valid, and test sets.
This code execution task requires models to maintain the values of multiple variables,
which has been previously shown to be a difficult task for Transformers
with only feedforward connections \citep{fan2020addressing}.

\subsection{Additional Results for Code Execution}
\label{app:lte_exp}
\paragraph{Token level print accuracy.}
First of all, as mentioned in the main text, the test accuracies reported in Table \ref{tab:exp_algo}
are on the sequence-level, i.e., an output sequence is counted as correct only if all output tokens in the sequence match the ground truth.
The sequence level accuracy is a good evaluation measure here since for most positions in the sequence (except at the end of \texttt{print} statement) the correct target is the no-output token.
This results in 0\% accuracy for the Linear Transformer, which might be shocking at first glance at Table \ref{tab:exp_algo}.
Thus, we also provide the token accuracies following the \texttt{print} statements.
The results can be found in Table \ref{tab:print_only}.
There we can see that the accuracies for the Linear Transformer are not zero:
above 20\% in both 3 and 5 variable cases. Nevertheless, they clearly underperform other models.

\begin{table}[h]
\caption{Token-level validation accuracies (\%) for the \textbf{print statements} on \textbf{code execution}.
Means and stds are computed with three seeds for 3-variable and six seeds for 5-variable cases.}
\label{tab:print_only}
\begin{center}
\begin{tabular}{lrr}
\toprule
&    \multicolumn{2}{c}{\# Variables}  \\ \cmidrule(r){2-3}  
&    \multicolumn{1}{c}{3} & \multicolumn{1}{c}{5}  \\
\midrule 
 LSTM        &    \mpm{\textbf{99.9}}{0.0}   & \mpm{\textbf{99.6}}{$\,\;$0.4} \\
Transformer            &  \mpm{98.6}{0.2} &  \mpm{75.5}{31.0}   \\
Linear Transformer      &  \mpm{24.6}{0.6}   &  \mpm{20.7}{$\,\;$1.4}  \\
Delta Net            & \mpm{99.5}{0.1} &    \mpm{97.2}{$\,\;$2.0}  \\ \midrule 
Delta RNN    &   \mpm{99.5}{0.0}   &  \mpm{\textbf{99.3}}{$\,\;$0.2}  \\
RDN  &   \mpm{\textbf{99.6}}{0.1} &   \mpm{98.6}{$\,\;$1.4} \\
\bottomrule
\end{tabular}
\end{center}
\end{table}

\paragraph{Model configurations.}
The Transformer architecture in Table \ref{tab:exp_algo} is
adopted from \citet{fan2020addressing}:
4 layers with a hidden dimension of 256 (where we use 16 heads instead of 4)
and a feedforward dimension of 1024, which yields 3.2\,M parameters (like for \citet{fan2020addressing}).
We use a dropout rate of 0.1.
The regular Transformer makes use of sinusoidal positional encoding (as is likely the case for \citet{fan2020addressing}) while all other models in Table \ref{tab:exp_algo} don't \citep{irie19:trafolm, schlag2021linear}.
All Transformer models use pre-activation
residual connections \citep{HeZRS16} and layer norm \citep{ba2016layer}.
Our LSTM model in Table \ref{tab:exp_algo} has one LSTM layer with a dimension of 256 and an input embedding of 128 which results in 405\,K parameters.
We train all models with a batch size of 64 using the Adam optimiser with a learning rate of 3e-4 for Transformer-family models and a learning rate of 3e-3 for the LSTM.
We clip the gradients in the LSTM model at 0.1. 
To train the regular Transformers, gradient accumulation was necessary to achieve the same batch size without hitting the GPU memory limit.
This was not the case for space efficient linear Transformer variants.
All models are trained for 200 epochs which takes no more than 23 hours for any model on a single 16\,GB P100 GPU.

\begin{table}[t]
\caption{Test accuracies (\%) on \textbf{code execution} with 5 variables.
Mean, standard deviation (std), the lowest (min) and highest (max) accuracies are computed over six runs.
The number of parameters (Prms.) is given in millions.}
\label{tab:exp_algo_deeper}
\begin{center}
\begin{tabular}{lccrrrr}
\toprule
& width & depth  & \mpm{mean}{std} & min & max & Prms. \\
\midrule
              
LSTM        &   256  & 1 &   93.2  $\pm \,\;$ {\small6.1} & 84.7 &  98.5 & 0.4 \\
                &   512  &   &  \textbf{97.7}  $\pm \,\;$ {\small1.1}  & 96.1 & 98.7 & 1.3 \\ \midrule
Delta Net        & 256 & 4   &   61.4 $\pm$ {\small 20.0}  & 26.2 &  85.7 & 3.2 \\   
Delta RNN    &  &    &   \textbf{85.1}  $\pm \,\;$ {\small 1.9} & 83.1 & 88.6 & 3.7 \\  
RDN  &  &  &    76.3  $\pm$ {\small 17.6} &  40.2 & 92.5 & 3.2 \\  
\midrule
Delta Net                     & 128  & 8      &  \mpm{62.7}{36.3} & 0.1 & 97.3 & 1.1 \\  
Delta RNN                     &  &  & \textbf{94.1} $\pm \,\;$ {\small2.7}   & 88.0 & 95.8 & 1.3 \\  
RDN                             & & & 85.0 $\pm \,\;$ {\small3.8} & 78.9 & 89.0& 1.1 \\  
\bottomrule
\end{tabular}
\end{center}

\end{table}

\paragraph{Model architecture ablation.}
Here we conduct a few additional experiments to understand the models' sensitivity to hyper-parameters.
We restrict our analysis to the setting with 5 variables in which the performance gap between models is large (Table \ref{tab:exp_algo}).
We train deeper but thinner models with 8 layers: each with a hidden size of 128 using 8 heads and a feed-forward dimension of 256. 
This yields a total of 1.1\,M parameters for all Transformer models except for the Delta RNN which has 1.3\,M parameters.
These deeper but thinner models can be trained within 10 hours using a single 16\,GB V100 GPU.
We present the results in the bottom part of Table \ref{tab:exp_algo_deeper}.
We don't report the performance of the regular Transformer since the 8-layer variant learns very slowly and does not improve over the initial 0\% sequence-level accuracy within 200 epochs of training after which we report the performance for all models\footnote{
Extra experiments with this 8-layer regular Transformer show that after 800 epochs with a dropout rate of 0.3, a test accuracy of $89.1\pm2.2$\% is achieved.
This is still worse than the performance of Delta RNN trained for 200 epochs, although the comparison is not even fair due to the longer training and extra tuning.}.

First, we observe that the Delta RNN with 8 layers can now match the performance of the baseline LSTM with 256 nodes.
However, increasing the LSTM hidden size to 512 (which gives a parameter count of 1.3\,M; equal to the Delta RNN's) further improves the LSTM.
Second, the Delta Net still remains unstable.
We tried several tricks to stabilise Transformers on algorithmic tasks \citep{csordas2021devil},
e.g. embedding initialisation and scaling, but with little success.
The problem seems intrinsically difficult for Transformer models, though we note that one of six runs  achieved a very good performance of 97.3\%.
Finally, we observe that the Recurrent Delta Net becomes more stable and performs better with a deep architecture. 

\subsection{Task Details for Sequential ListOps}
\label{app:listops}
The ListOps task \citep{nangia-bowman-2018-listops} consists of list operation execution
which is a typical test for hierarchical structure learning.
A list is constructed using elementary list-operations written in prefix notation
(typically one of six operations: maximum, minimum, median followed by floor operation,
sum modulo 10, first and last element retrieval)
with a random number of random arguments chosen to be either a single digit integer or a sub-list
which itself has random arguments.
While early research comparing self-attention to RNNs \citep{TranBM18} has shown
some advantages of recurrence in hierarchical structure learning,
more recent work \citep{lu2021pretrained}
reports Transformers to also outperform LSTMs on ListOps.
Also relevant here, \citet{tay2020long} report linear Transformer variants (Linear Transformers and Performers) to underperform other Transformer variants by a large margin on ListOps.
It is thus natural to evaluate our models on this task as models at the intersection of recurrent and self-attention based models.
We construct a simple variant of ListOps which only makes  use of maximum \texttt{MAX}, minimum \texttt{MIN}, and first element retrieval \texttt{FIRST} operations. This turns out to be hard enough to shed light on the differences between our models.
By construction, the targets are single digit integers.
The number of arguments in each list or sub-list is random but less than the pre-determined maximum number (here set to five, following \citet{nangia-bowman-2018-listops})
and we control the difficulty of the task by changing the problem depth.
Here is a depth-two example:

\begin{table}[h]
\small		
\label{tab:example_listops}
\begin{center}
\setlength{\tabcolsep}{0.5em}
\begin{tabular}{lr}
In: &  \texttt{[MAX 6 1 [FIRST 2 3 ] 0 [MIN 4 7 1] ]} \\
Out: & 6 \\
\end{tabular}
\end{center}
\end{table}
In our setting, the task with depth 10 only contains sequences with depth 10\footnote{
However, here the \textit{depth} is 
simply defined as the depth of nested operations.
Since the used operations do not always have to evaluate all arguments to obtain the result, the \textit{effective computation} may be shallower.
This problem has been addressed in
a better version of ListOps in
our more recent work \citep{csordas2021neural}.
}.
Here, we refer to the task as ``sequential ListOps'',
as we let the model read the sequence only once from left to right in an auto-regressive fashion.
As for the code execution experiments, we randomly generate 10,000 sequences for training and 1,000 sequences each for validation and test.
The lengths for the depth 10 case vary from 37 to 364 with an average length of 98 tokens.
For the depth 15 case, the lengths are between 61 and 676, with an average of about 190 tokens.
All experiments were conducted using a single 16\,GB P100 GPU.
We use the same experimental settings as in the code execution task, and
the experiments for depth 10 and 15 take less than 4 and 16 hours,
respectively.

\subsection{Ablation Study for the LSTM on Sequential ListOps}
\label{app:listops}
While the main goal of Table \ref{tab:exp_algo} (Sec.~\ref{sec:listops})
was to compare different fast weight programmer variants under the same model configurations,
we also pointed out that the performance of the baseline LSTM dramatically drops for the sequential ListOps task by increasing the list depth from 10 to 15.
In Sec.~\ref{sec:listops}, we hypothesised the reason for the performance
drop of the LSTM for the depth-15 case of sequential ListOps to be the small hidden size of the LSTM and the increase of sequence lengths in the depth-15 case.
Here we provide the corresponding ablation study.
Table \ref{tab:lstm_listops} shows the performance
of the LSTM with different hidden layer sizes.
We find that increasing the hidden size
effectively help the LSTM on this task.

\begin{table}[h]
\caption{Test accuracies (\%) with standard deviations over three runs for the \textbf{LSTM} on the \textbf{depth-15} case of \textbf{Sequential ListOps}.}
\label{tab:lstm_listops}
\begin{center}
\begin{tabular}{rr}
\toprule
Hidden size &    \multicolumn{1}{c}{Mean accuracy $\pm$ std} \\
\midrule 
 256      & 24.4 $\pm \,\;$ {\small 1.1} \\
1,024      & 24.4 $\pm \,\;$ {\small 0.7} \\
2,048      &  \mpm{35.9}{13.0} \\
4,096 & \textbf{72.2} $\pm \,\;$ {\small 1.6} \\
\bottomrule
\end{tabular}
\end{center}
\end{table}

\section{Experimental Details and Additional Results for RL in Atari 2600}
\label{app:rl}
\paragraph{Settings.}
We use the \texttt{polybeast} implementation from \texttt{Torchbeast} \citep{kuttler2019torchbeast} with modifications limited to model architectures.
We train all our models using RMSProp \citep{rmsprop} with a learning rate of 0.0006, an epsilon of 0.01, and gradient clipping at 40.
We use entropy regularisation with a weight of 0.01.
The backpropagation span is 50 and the batch size is 32.
The model architecture and evaluation method is described in the main text.
All Transformer variants make use of pre-activation
residual connections \citep{HeZRS16, parisotto2020stabilizing} and layer norm \citep{ba2016layer}.
The number of actors for IMPALA is 48.
No action repeat is used.
No time limit is set for evaluation.
Rewards are clipped between -1 and 1.
The OpenAI Gym implementation of the Atari learning environment \citep{machado2018revisiting} is used.
The only source of stochasticity is the default sticky action.
We train expert models using the game specific action spaces (models for \textit{Amidar} and \textit{James Bond} were trained with an action space size of 6, which is smaller than the full action space but enough to play these games).
We train on 2 GPUs (either 16\,GB P100 or 32\,GB V100).
An experiment for one game takes about 1.5 days.
Evaluation is done at 50\,M and 200\,M environmental steps, which are reported in Table \ref{tab:atari_50m} and \ref{tab:atari_200m}.
For cases where performance did not improve after 50\,M and 200\,M, we report the performance at 50\,M again in Table \ref{tab:atari_200m} (we experienced this for \textit{Bank Heist} and \textit{Robotank}; for \textit{Pong} 50~M steps are enough to consistently achieve the perfect score).

In what follows, we provide additional model comparisons.

\begin{figure}[h]
    \begin{minipage}{0.45\textwidth}
    \begin{center}
        \includegraphics[width=1.05\columnwidth]{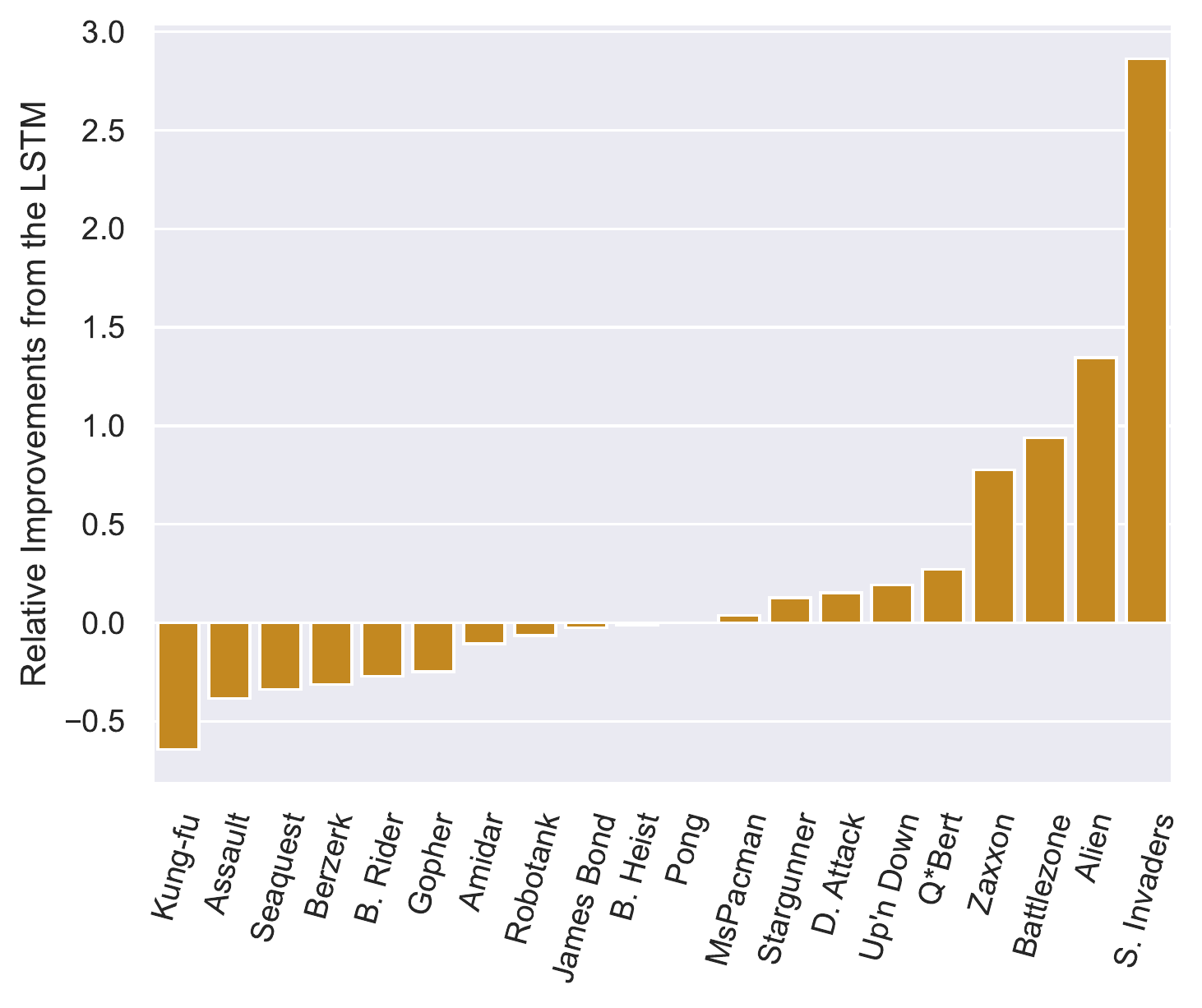}
        \caption{Rel.~improvements in test scores obtained by the \textbf{feedforward baseline} compared to LSTM after \textbf{50\,M} env.~steps.}
        \label{fig:atari_ff_50m}
    \end{center}
\end{minipage}
\hspace{8mm}
    \begin{minipage}{0.45\textwidth}
    \begin{center}
        \includegraphics[width=1.05\columnwidth]{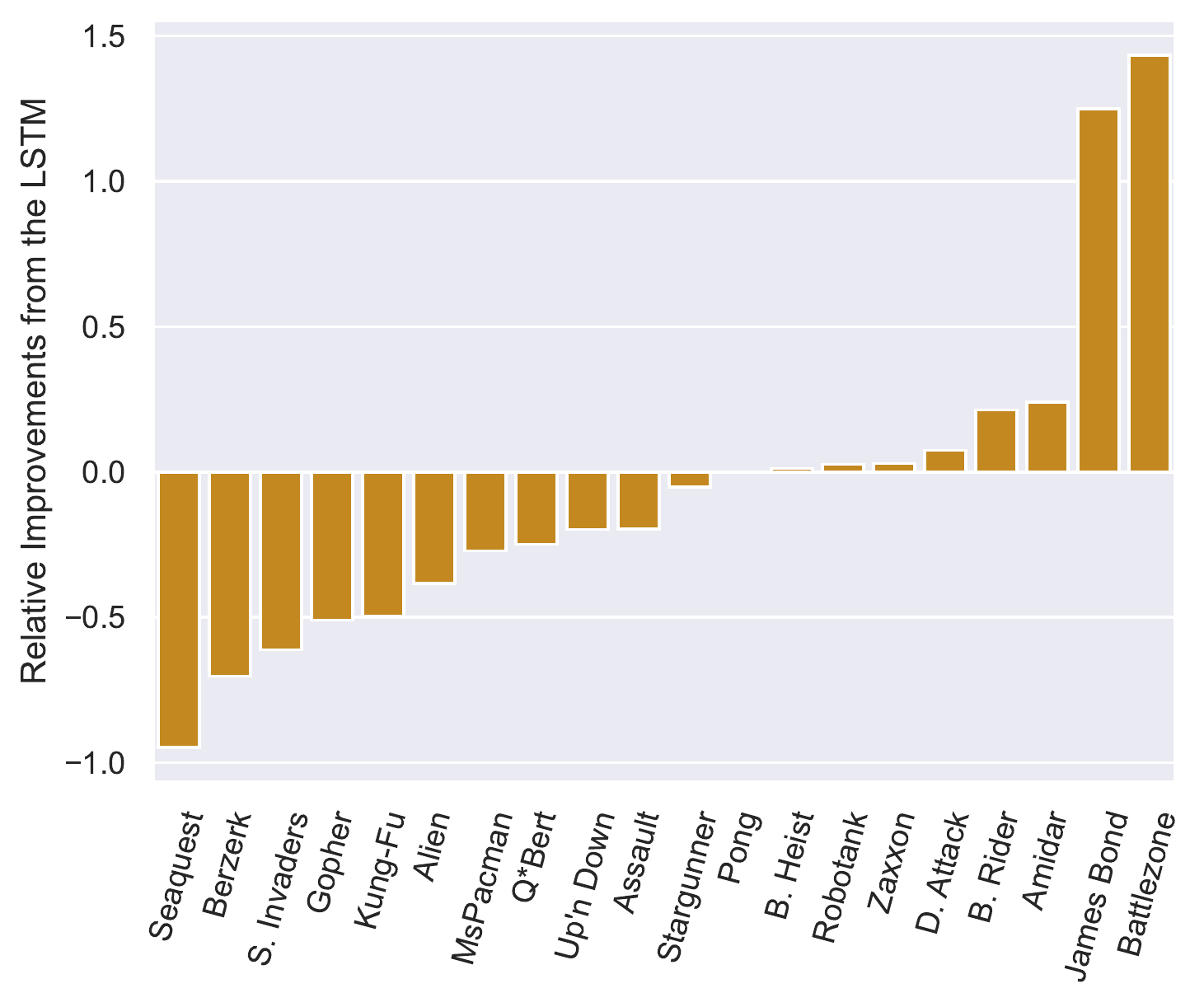}
        \caption{Rel.~improvements in test scores obtained by the \textbf{feedforward baseline} compared to LSTM after \textbf{200\,M} env.~steps.}
        \label{fig:atari_ff_200m}
    \end{center}
\end{minipage}
\end{figure}

\paragraph{Feedforward vs.~LSTM.}
On Atari, models without recurrence are also known to perform well in many
environments \citep{MottZCWR19}.
Since it is not easy to compare RL systems across different settings \citep{andrychowicz2021matters}, we train our own feedforward baseline.
The feedforward baseline is simply obtained by removing the LSTM layer in the LSTM model, which corresponds to the model of \citet{EspeholtSMSMWDF18}.
At 50\,M steps (Figure \ref{fig:atari_ff_50m}; \textit{orange}), there are 8 games in which the feedforward baseline clearly outperforms the LSTM, and in 8 other games the trend is reversed.
At 200\,M steps (Figure \ref{fig:atari_ff_200m}; \textit{orange}), the LSTM performs clearly better in 10 games, whereas the feedforward net
clearly dominates only in 4 games.

\paragraph{Feedforward baseline with more parameters.}
In the comparison above,
the LSTM baseline has 1.6\,M parameters, more than the 1.1\,M parameters of the feedforward baseline (while we note that the RDN has slightly fewer parameters than the LSTM, namely, 1.5 M).
To verify that the improvements obtained by the LSTM
are not due to the increased parameter count,
we build a larger feedforward baseline with 1.7 M parameters by replacing the LSTM layer by one feedforward highway-gated layer \citep{srivastava2015icml} (to keep it as similar as possible to the LSTM baseline).
Here the output from the vision stem is first projected
to a 320-dimensional vector
which is followed by a 256-dimensional highway-gated layer.
We evaluate this model on four environments on which the LSTM outperforms the 1.1 M-param feedforward baseline.
Table \ref{tab:atari_ff_larger} shows
the corresponding results.
The extra parameters yield improvements only on
\textit{S. Invader}, without matching LSTM's  performance.
So we can confirm that the dominance of LSTM over feedforward models in these games  is not simply due to the higher parameter count.

\begin{table}[h]
\caption{Performance of feedforward baseline with more parameters.
}
\label{tab:atari_ff_larger}
\begin{center}
\begin{small}
\begin{tabular}{lrrrrr}
\toprule
   & Params. & \multicolumn{1}{c}{Berzerk} &  \multicolumn{1}{c}{Gopher}  &  \multicolumn{1}{c}{Seaquest}  & \multicolumn{1}{c}{S. Invader}  \\ \midrule
LSTM   &  1.5\,M    &  \mpm{1,150}{92}  &  \mpm{124,914}{22,422}   & \mpm{12,643}{1,627}  &  
137,657 $\pm \,\;$ 2,276
\\ \midrule

FF   &  1.1\,M   &  \mpm{\textbf{343}}{23}   &  \textbf{61,350} $\pm \,\;$ 3,891  & \textbf{667} $\pm \,\;\,\;\;\;\,$ 1   & 53,455 $\pm \,\;$ 6,694 \\

FF gated  & 1.7\,M  &  \mpm{320}{29} & 42,851 $\pm \,\;$ 7,653    & 660 $\pm \,\;\,\;\;\;\,$ 0    & \mpm{\textbf{95,629}}{11,991}  \\
\bottomrule
\end{tabular}
 \end{small}
\end{center}
\end{table}

\begin{figure}[h]
    \begin{minipage}{0.45\textwidth}
    \begin{center}
        \includegraphics[width=1.05\columnwidth]{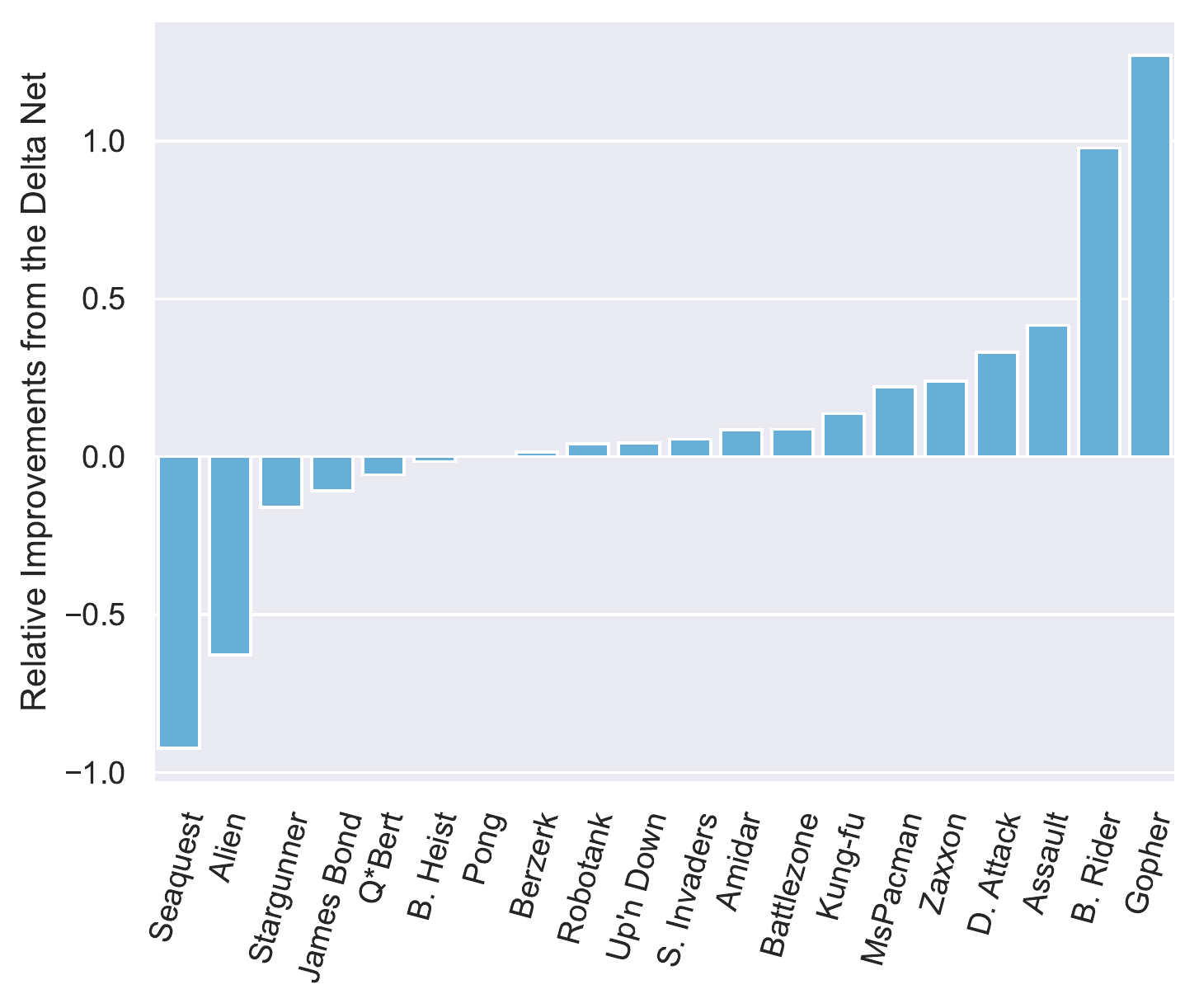}
        \caption{Rel.~improvements in test scores obtained by the \textbf{Recurrent Delta Net} compared to the \textbf{Delta Net} after \textbf{50\,M} env.~steps.}
        \label{fig:atari_delta_50m}
    \end{center}
\end{minipage}
\hspace{8mm}
    \begin{minipage}{0.45\textwidth}
    \begin{center}
        \includegraphics[width=1.05\columnwidth]{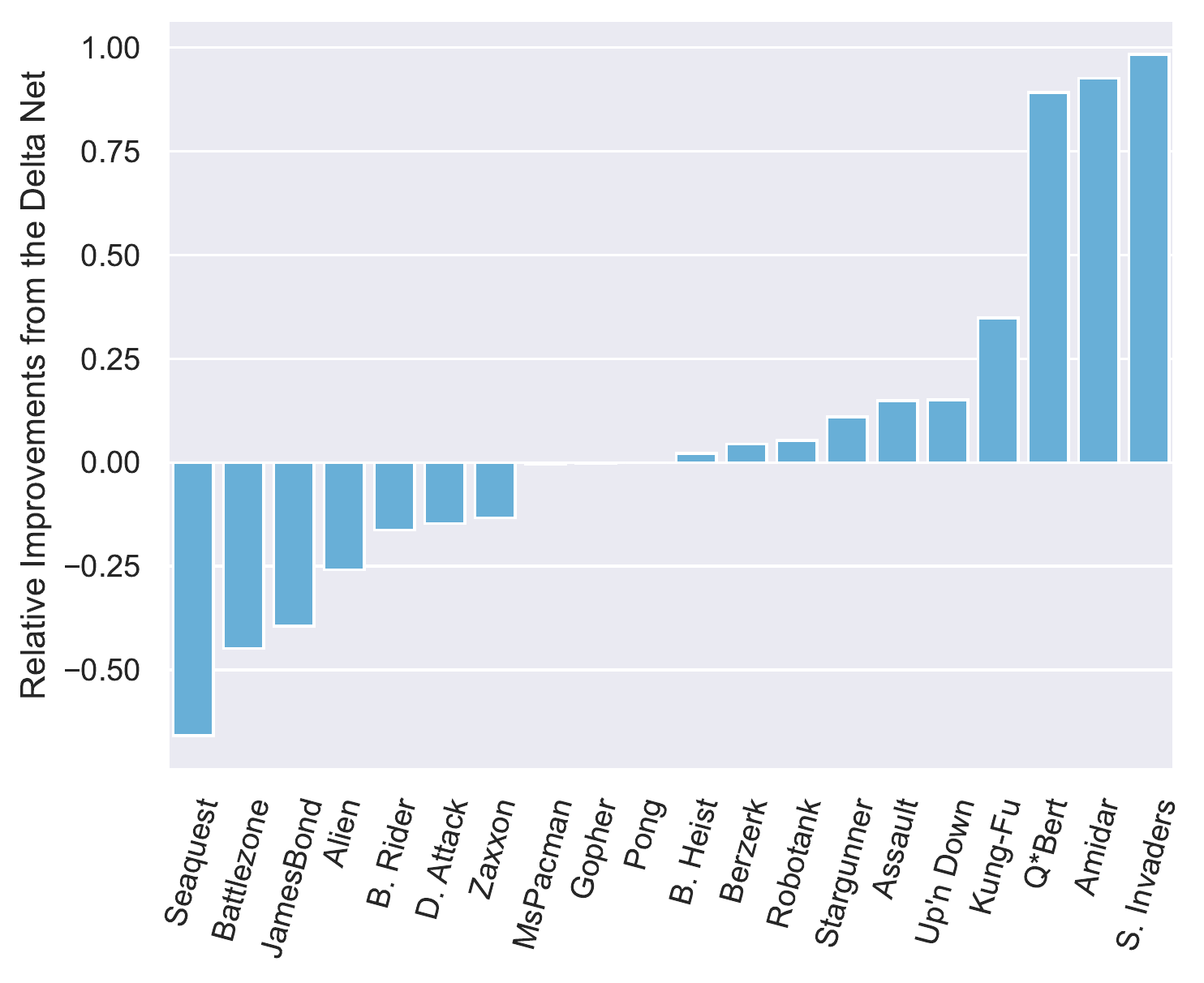}
        \caption{Rel.~improvements in test scores obtained by the \textbf{Recurrent Delta Net} compared to the \textbf{Delta Net} after \textbf{200\,M} env.~steps.}
        \label{fig:atari_delta_200m}
    \end{center}
\end{minipage}
\end{figure}

\paragraph{Recurrent Delta Net vs.~Delta Net.}
We also compare the Recurrent Delta Net to a stronger baseline, the Delta Net.
The results are shown in Figures \ref{fig:atari_delta_50m} and \ref{fig:atari_delta_200m} (\textit{sky blue}).
While the RDN performs equally well or better than the baseline Delta Net on 13 games at 200~M steps, there are also 7 games where the Delta Net is better.
We thus can not guarantee strict benefits of  additional recurrence here.
Again we note that compared to other models, both the Delta Net and Recurrent Delta Net achieve outstanding performance on \textit{Q*Bert}.

\begin{figure}[h]
    \begin{minipage}{0.45\textwidth}
    \begin{center}
        \includegraphics[width=1.05\columnwidth]{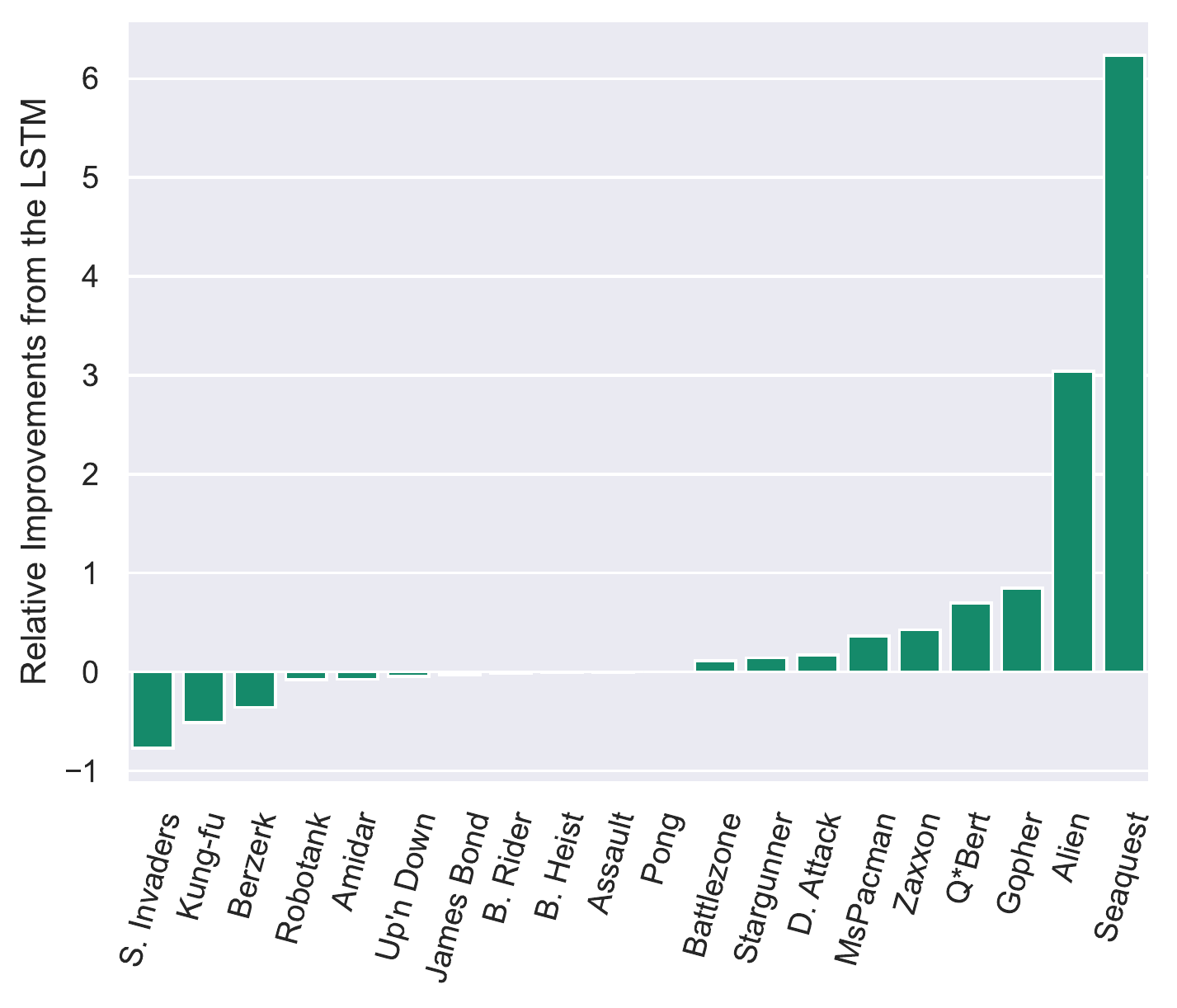}
        \caption{Rel.~improvements in test scores obtained by the \textbf{Delta RNN} compared to the \textbf{LSTM} after \textbf{50\,M} env.~steps.}
        \label{fig:atari_rnn_50m}
    \end{center}
\end{minipage}
\hspace{8mm}
    \begin{minipage}{0.45\textwidth}
    \begin{center}
        \includegraphics[width=1.05\columnwidth]{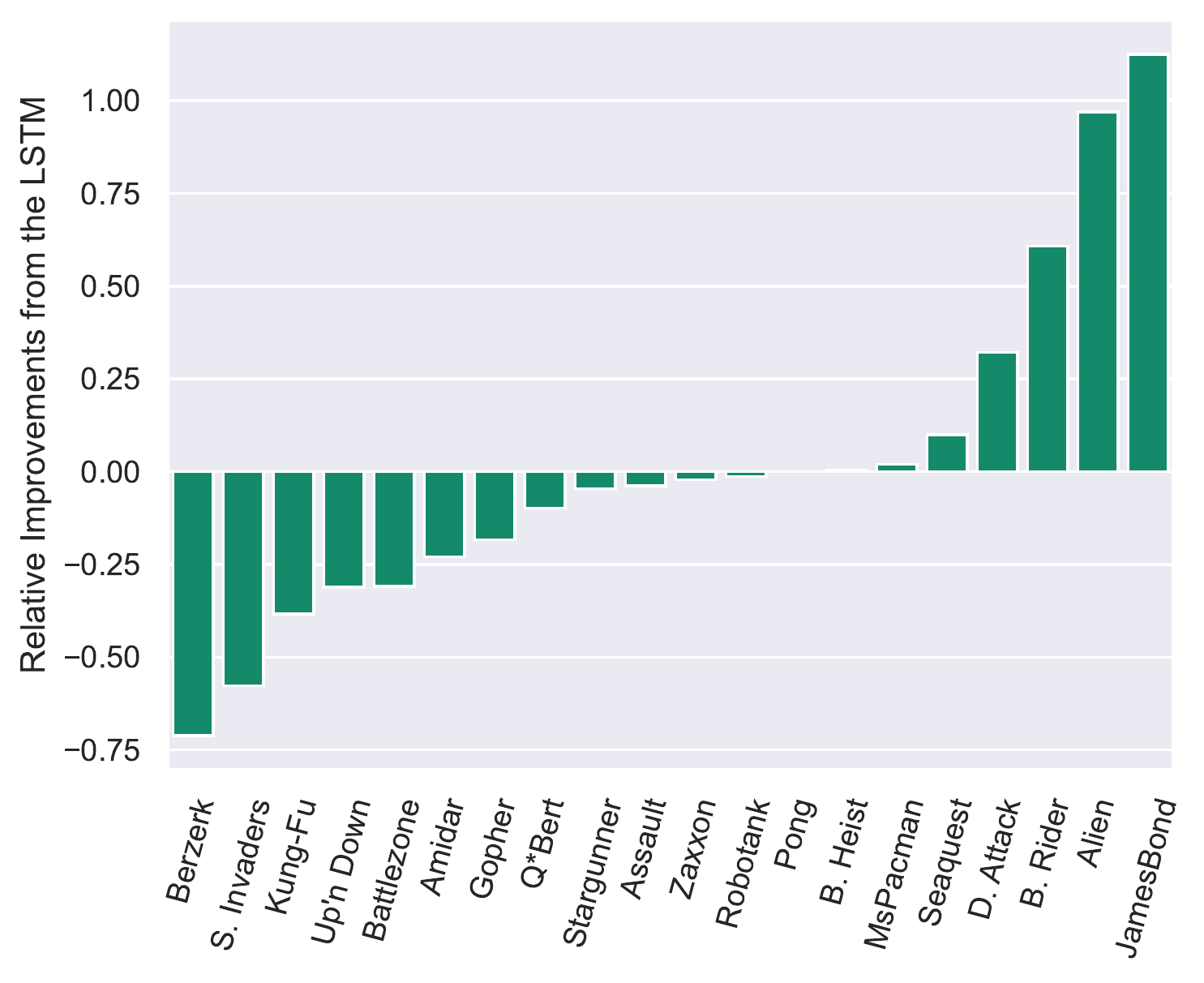}
        \caption{Rel.~improvements in test scores obtained by the \textbf{Delta RNN} compared to the \textbf{LSTM} after \textbf{200\,M} env.~steps.}
        \label{fig:atari_rnn_200m}
    \end{center}
\end{minipage}
\end{figure}

\paragraph{Delta RNN vs.~LSTM.}
We also evaluate the Delta RNN (Sec.~\ref{sec:fast-net}) in this RL setting.
We first compare it to the LSTM baseline.
As shown in Figures \ref{fig:atari_rnn_50m} and \ref{fig:atari_rnn_200m} (\textit{green}), the Delta RNN clearly outperforms the LSTM on a few games at 50~M steps.
However, the performance gaps reduce across all games after 200~M steps.
Overall, the performance is close in 7 games, in favour of the LSTM in 8 games, and in favour of the Delta RNN in 5 games.

\begin{figure}[h]
    \begin{minipage}{0.45\textwidth}
    \begin{center}
        \includegraphics[width=1.05\columnwidth]{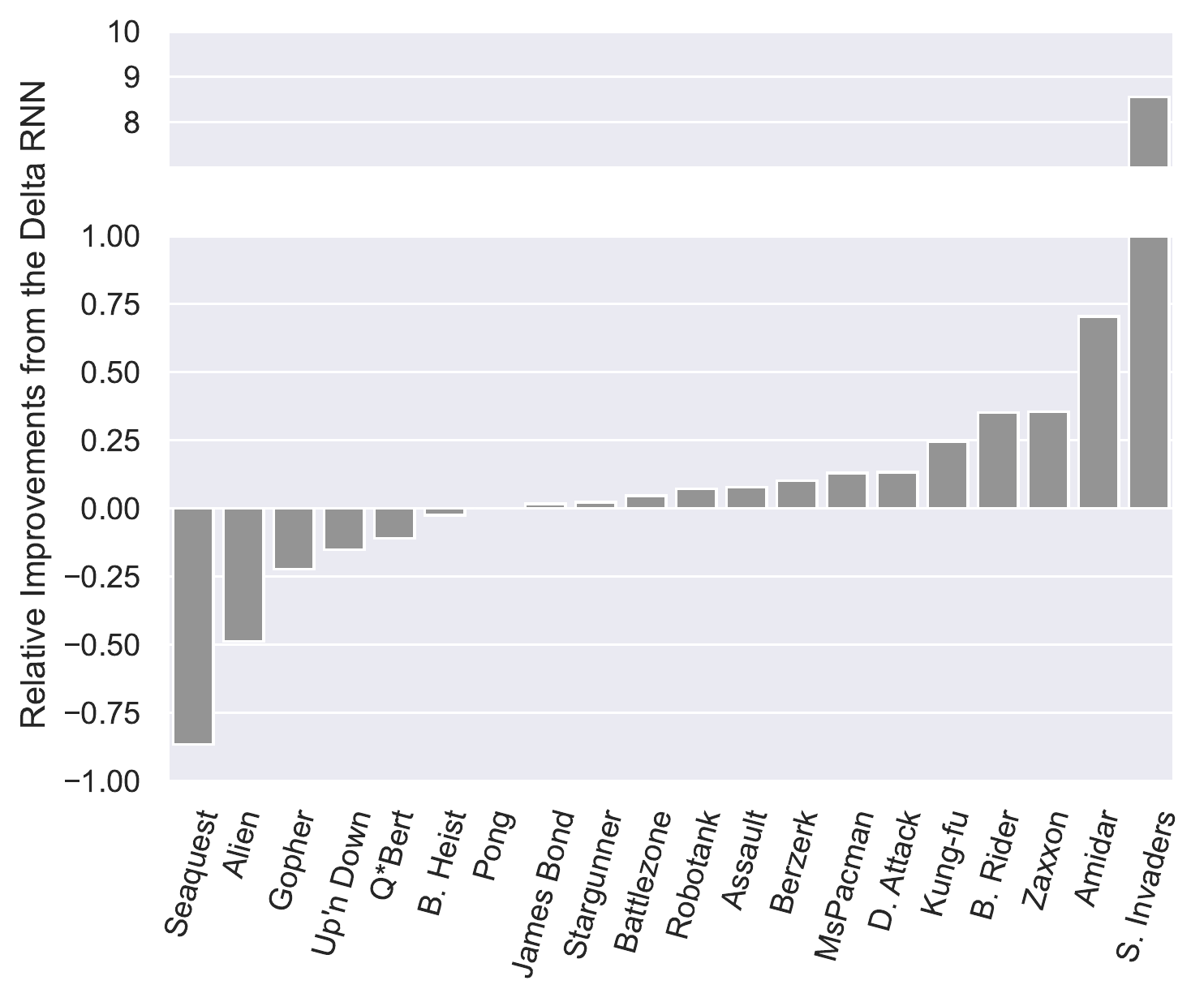}
        \caption{Rel.~improvements in test scores obtained by the \textbf{Recurrent Delta Net} compared to the \textbf{Delta RNN} after \textbf{50\,M} env.~steps.}
        \label{fig:atari_vs_delta_rnn_50m}
    \end{center}
\end{minipage}
\hspace{8mm}
    \begin{minipage}{0.45\textwidth}
    \begin{center}
        \includegraphics[width=1.05\columnwidth]{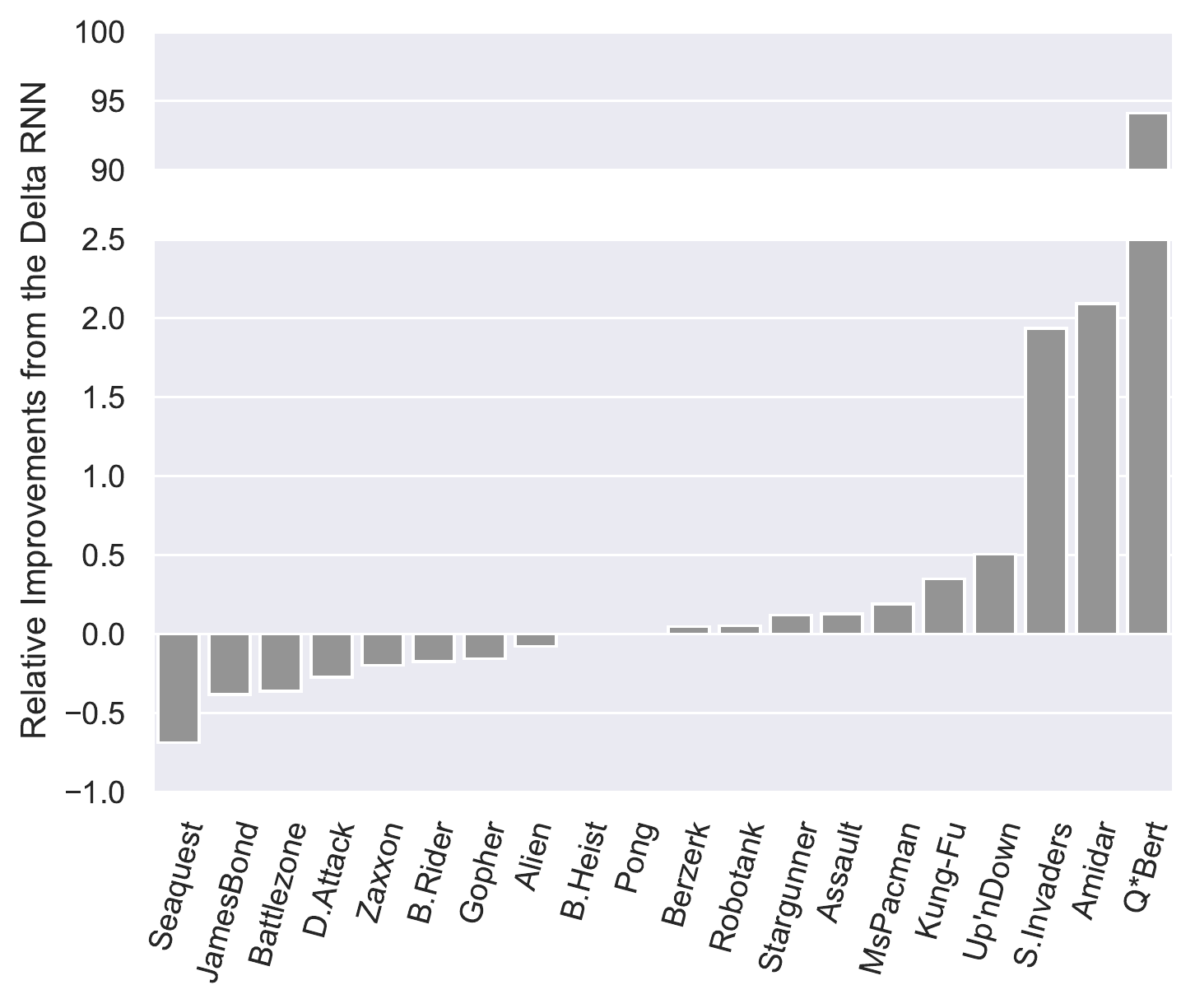}
        \caption{Rel.~improvements in test scores obtained by the \textbf{Recurrent Delta Net} compared to the \textbf{Delta RNN} after \textbf{200\,M} env.~steps.}
        \label{fig:atari_vs_delta_rnn_200m}
    \end{center}
\end{minipage}
\end{figure}

\paragraph{Recurrent Delta Net vs.~Delta RNN.}
Finally, we also compare the Recurrent Delta Net to the Delta RNN.
Figures \ref{fig:atari_vs_delta_rnn_50m} and \ref{fig:atari_vs_delta_rnn_200m} (\textit{grey}) present our results.
In 16 games, the relative performance gap is within 50\%.
In one game (\textit{Seaquest}), the Delta RNN outperforms the RDN. In 3 games, the RDN clearly outperforms the Delta RNN at 200~M steps. 

Overall, the Recurrent Delta Net tends to yield decent performance
compared to all baselines.
While the performance gaps between the Recurrent Delta Net
and the Delta RNN are rather close, the Recurrent Delta Net
performs particularly well in a few games.
As mentioned in the main text, trying deeper architectures
might be a straight-forward way to obtain better scores.

\section{Comments on Nomenclature}
To simplify references to specific Fast Weight Programmers,
we gave short names to all of them, such as Delta RNN or Recurrent Delta Net. We did not cover,
however, many other possible combinations of slow and fast networks as well as update rules (which are the elementary programming instructions of FWPs).
This calls for a systematic nomenclature to specify the various FWP  types.  For a given FWP, 
one could use ``\textit{slow-net/update-rule}" as a \textit{prefix} and "\textit{fast-net}" architecture as a \textit{suffix}.
For example, the Delta RNN is an FWP with a fast RNN and a feedforward slow net using the delta rule as elementary programming instruction.
Therefore, using the convention above, the full name of the Delta RNN would be ``\textit{Feedforward/Delta fast RNN}."
The full name of the Recurrent Delta Net would be ``\textit{Recurrent/Delta fast Linear Net}," and so on.
This is also compatible with the baseline Delta Net, whose full name would be ``\textit{Feedforward/Delta fast Linear Net}."

%
%

%

%

\newcommand{\mpms}[2]{#1 $\pm$ {\scriptsize #2}}  


\begin{table}[h]
\scriptsize	
\caption{Performance after \textbf{50\,M} environmental steps.
Reported scores are mean and std of 5 mean-scores obtained over 30 episodes (total of 150 different test episodes).
We remind the reader that we denote the Linear Transformer \citep{katharopoulos2020transformers} as LT, and our Recurrent Delta Network as RDN.
The numbers of parameters are: 1.1~M for the feedforward model,
1.6~M for the LSTM, 1.5~M for the Linear Transformer, the Delta Net, and the Recurrent
Delta Net, and finally 1.6\,M for the Delta RNN.
}
\label{tab:atari_50m}
\begin{center}
\setlength{\tabcolsep}{0.6em}
\begin{tabular}{lrrrrrr}
\toprule
           & Feedforward & LSTM & LT  & Delta Net & RDN & Delta RNN  \\ \midrule
Alien    & \mpms{1,985}{90} & \mpms{846}{81} & \mpms{2,135}{184} & \mpms{\textbf{4,704}}{452} & \mpms{1,754}{48} & \mpms{3,420}{834} \\   
Amidar     &  \mpms{208}{11} & \mpms{233}{10}  & \mpms{320}{16} & \mpms{339}{28} &  \mpms{\textbf{368}}{23} & \mpms{216}{14} \\
Assault     & \mpms{4,658}{2,147} & \mpms{7,551}{1,774} & \mpms{2,764}{380} & \mpms{5,710}{2,643} & \mpms{\textbf{8,088}}{2,851} & \mpms{7,503}{2,794}  \\  
Battlezone & \mpms{\textbf{12,267}}{620} & \mpms{6,327}{380} & \mpms{933}{351} & \mpms{6,780}{461}&  \mpms{7,373}{431}  & \mpms{7,040}{1,098} \\
Berzerk    & \mpms{326}{21} &  \mpms{\textbf{474}}{17} & \mpms{323}{6} & \mpms{331}{24} & \mpms{336}{27} &  \mpms{305}{8} \\   
B. Heist   & \mpms{323}{13} & \mpms{\textbf{327}}{11} & \mpms{309}{11} & \mpms{321}{8} & \mpms{316}{10} & \mpms{324}{10}  \\ 
B.~Rider  & \mpms{9,932}{1,592} & \mpms{13,638}{1,571} & \mpms{6,695}{941} & \mpms{9,185}{630} & \mpms{\textbf{18,156}}{1,522} & \mpms{13,429}{884} \\  
D. Attack & \mpms{36,255}{3,566} & \mpms{31,447}{1,850} & \mpms{8,939}{950} & \mpms{31,359}{3,362} & \mpms{\textbf{41,726}}{6,308} & \mpms{36,807}{3,700}\\  
Gopher & \mpms{10,356}{378} & \mpms{13,765}{808} & \mpms{8,197}{1,720} & \mpms{8,707}{2,381} & \mpms{19,775}{1,448}  & \mpms{\textbf{25,445}}{1,963}  \\
James Bond          & \mpms{2,942}{56} & \mpms{3,020}{68} & \mpms{2,425}{174} &\mpms{\textbf{3,338}}{137} & \mpms{2,979}{176} & \mpms{2,929.3}{408} \\  
 Kung-fu    & \mpms{5,449}{82} & \mpms{\textbf{15,216}}{818} & \mpms{3,722}{330} & \mpms{8,095}{240} & \mpms{9,201}{384} &  \mpms{7,388}{491}  \\
MsPacman     & \mpms{1,737}{53} & \mpms{1,676}{86} & \mpms{1,647}{101} & \mpms{2,116}{30} & \mpms{\textbf{2,584}}{121} &  \mpms{2,287}{32} \\  
Pong           & \mpms{21}{0}  & \mpms{21}{0}  & \mpms{21}{0}   & \mpms{21}{0}   & \mpms{21}{0}  & \mpms{21}{0}      \\
Q*Bert       &  \mpms{4,967}{266} & \mpms{3,905}{252}  & \mpms{4,693}{195} & \mpms{6,248}{204} & \mpms{5,897}{357} &  \mpms{\textbf{6,626}}{240}\\  
Robotank    & \mpms{7.1}{0.7} & \mpms{\textbf{7.6}}{0.7} & \mpms{4.8}{0.3} & \mpms{7.2}{0.7} & \mpms{7.5}{0.8} & \mpms{7.0}{0.5}  \\  
Seaquest    & \mpms{469}{1} & \mpms{708}{1} & \mpms{1,812}{61} & \mpms{\textbf{8,853}}{937} & \mpms{686}{1} &   \mpms{5,123}{335} \\
S.~Invaders  & \mpms{\textbf{48,150}}{7,233} & \mpms{12,461}{1,624} & \mpms{2,345}{74} & \mpms{25,769}{10,156} & \mpms{27,213}{3,359} & \mpms{2,847}{10} \\
Stargunner  & \mpms{9,397}{2,193} & \mpms{8,337}{1,094} & \mpms{8,915}{713} & \mpms{\textbf{11,599}}{3,454} &  \mpms{9,737}{1,396}     & \mpms{9,523}{2,214}  \\  
Up'n down   & \mpms{\textbf{185,632}}{16,490} & \mpms{155,847}{15,318} & \mpms{57,435}{2,283} & \mpms{120,806}{16,261} & \mpms{126,140}{19,078} &  \mpms{148,759}{28,492} \\
Zaxxon & \mpms{4863}{872} & \mpms{2,737}{121} & \mpms{2,719}{701} & \mpms{4,265}{263} & \mpms{\textbf{5,285}}{504}  & \mpms{3,903}{648}  \\   %
\bottomrule
\end{tabular}
\end{center}
\end{table}


\begin{table}[h]
\scriptsize
\caption{Performance after \textbf{200\,M} environment steps.
Reported scores are mean and std of 5 mean-scores obtained over 30 episodes (total of 150 different test episodes).
We remind the reader that we denote the Linear Transformer \citep{katharopoulos2020transformers} as LT, and our Recurrent Delta Network as RDN.
In cases where performance did not improve after 50\,M, we report the
performance at 50\,M.
}
\label{tab:atari_200m}
\begin{center}
\setlength{\tabcolsep}{0.6em}
\begin{tabular}{lrrrrrr}
\toprule
         & Feedforward  & LSTM & LT & Delta Net &  RDN & Delta RNN \\ \midrule
Alien     & \mpms{3,816}{139} & \mpms{6,184}{558}  & \mpms{4,751}{530} & \mpms{\textbf{15,133}}{1,122} &  \mpms{11,220}{621} & \mpms{12,177} {968} \\   
Amidar     &  \mpms{433}{27}   & \mpms{349}{22} & \mpms{646}{32} & \mpms{432}{27} &  \mpms{\textbf{832}}{11} & \mpms{269}{17} \\
Assault      & \mpms{6,407}{3,430} & \mpms{7,977}{2,611}  & \mpms{6,465}{1,437} & \mpms{7,525}{1,703} &  \mpms{\textbf{8,647}}{3,061} & \mpms{7,670}{952} \\  
Battlezone & \mpms{60,527}{12,345} & \mpms{\textbf{24,873}}{1,240} & \mpms{2,667}{386} & \mpms{19,907}{1,409} &  \mpms{10,980}{1,104} & \mpms{17,180}{1,493} \\  
Berzerk    & \mpms{343}{23}  & \mpms{\textbf{1,150}}{92} & \mpms{480}{38} & \mpms{333}{7} & \mpms{348}{17} & \mpms{332}{17}\\   
B. Heist   & \mpms{\textbf{331}}{10} & \mpms{327}{11} & \mpms{317}{8} & \mpms{321}{8} &  \mpms{328}{10} & \mpms{328}{8} \\  
B.~Rider   &\mpms{21,873}{2,000} & \mpms{18,024}{933} & \mpms{22,444}{755} & \mpms{28,594}{5,508} & \mpms{23,934}{2,292} & \mpms{\textbf{28,973}}{3,663} \\  
D. Attack  & \mpms{74,904}{10,941} & \mpms{69,750}{9,593} & \mpms{57,715}{5,009} & \mpms{78,601}{16,907} & \mpms{67,039}{5,714} & \mpms{\textbf{92,205}}{17,933}\\  
Gopher    & \mpms{61,350}{3,891} & \mpms{\textbf{124,914}}{22,422} & \mpms{48,261}{7,727} & \mpms{86,168}{5,069} &  \mpms{86,008}{11,815} &  \mpms{101,974}{10,200} \\  
James Bond            & \mpms{\textbf{56,459}}{7,292}  & \mpms{25,106}{5,889} & \mpms{16,223}{1,118} & \mpms{54,336}{7,165} &  \mpms{32,923}{7,968} & \mpms{53,344}{4,768} \\  
Kung-fu      & \mpms{12,292}{613} & \mpms{\textbf{24,447}}{407} & \mpms{13,969}{803} & \mpms{15,064}{929} &  \mpms{20,319}{363} & \mpms{15,068}{513}  \\  
MsPacman       & \mpms{2,499}{141} & \mpms{3,431}{197} & \mpms{3,052}{128} & \mpms{\textbf{4,180}}{139} &  \mpms{4,168}{585} &  \mpms{3,500}{205} \\  
Pong           & \mpms{21}{0}  & \mpms{21}{0}  & \mpms{21}{0}   & \mpms{21}{0}   &  \mpms{21}{0}  & \mpms{21}{0}      \\
Q*Bert          & \mpms{8,655}{371} &  \mpms{11,513}{910} & \mpms{8,389}{349} &  \mpms{521,839}{36,192} & \mpms{\textbf{987,275}}{0} & \mpms{10,381}{1,259} \\  
Robotank     &  \mpms{7.8}{0.8} &  \mpms{7.6}{0.7} & \mpms{7.7}{0.9} & \mpms{7.5}{0.4} &  \mpms{\textbf{7.9}}{0.6} &  \mpms{7.5}{0.5} \\  
Seaquest    & \mpms{667}{1} & \mpms{12,643}{1,627} & \mpms{12,425}{1,910} & \mpms{12,790}{1,512} &  \mpms{4,373}{504} & \mpms{\textbf{13,898}}{1,674}\\  
S.~Invaders    & \mpms{53,455}{6,694} & \mpms{137,657}{2,276} & \mpms{2,333}{110} & \mpms{86,132}{5,483} &  \mpms{\textbf{170,871}}{80}  & \mpms{58,181}{14,987}\\
Stargunner  & \mpms{11,564}{4,598}  & \mpms{12,194}{7,038} & \mpms{12,035}{6,995} & \mpms{11,734}{6,827} & \mpms{\textbf{13,026}}{6,431} & \mpms{11,635}{6,065} \\  
Up'n down    & \mpms{185,632}{16,490} & \mpms{231,157}{10,603} & \mpms{\textbf{252,555}}{16,331} &  \mpms{208,563}{22,803}&  \mpms{240,003}{26,849} &  \mpms{159,296}{25,013} \\
Zaxxon & \mpms{\textbf{11,960}}{538} & \mpms{11,619}{663} & \mpms{7,371}{932} & \mpms{1,0523}{568} &  \mpms{9,126}{313} & \mpms{11,365}{678} \\   
\bottomrule
\end{tabular}
\end{center}

\end{table}

\end{document}